\documentclass{article}

\usepackage[final]{neurips_2025}

\makeatletter
\renewcommand{\@noticestring}{%
  39th Conference on Neural Information Processing Systems
  (NeurIPS 2025), PersonaLLM Workshop.
}
\makeatother

\usepackage[utf8]{inputenc} 
\usepackage[T1]{fontenc}    
\usepackage{hyperref}       
\usepackage{url}            
\usepackage{booktabs}       
\usepackage{amsfonts}       
\usepackage{nicefrac}       
\usepackage{microtype}      
\usepackage{xcolor}         

\usepackage{color}
\usepackage{algorithm}
\usepackage{algorithmic}
\usepackage{graphicx}
\usepackage{amsmath}
\usepackage{amsfonts}
\usepackage{amssymb}
\usepackage{amsthm}
\usepackage{mathtools}
\usepackage{multirow}
\usepackage{balance} 
\usepackage{mathrsfs}
\usepackage{makecell}
\usepackage{booktabs,array}
\usepackage[most]{tcolorbox}

\setcitestyle{square}

\newcommand{\ModelName}{{SCorPion}}

\title{\protect\raisebox{-0.5ex}{\protect\includegraphics[height=3ex]{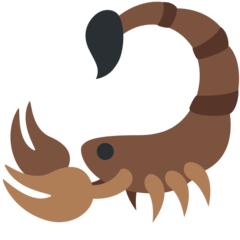}}  SCorPion: A Strategic, Controllable and Personified Conversational Companion}

\author{%
  \textbf{Luo Ji \quad Xiyuan Chen \quad Dan Wang \quad Qingqing Gu \quad Teng Chen} \\
  \textbf{Feixiang Guo \quad Yetao Wu \quad Jinxian Qu \quad Yue Zhao \quad Kun Fan} \\
  \textbf{Ningyuan Xi \quad Hange Zhou \quad Zhonglin Jiang \quad Yong Chen \quad Hongyan Li} \\
  \\
  Geely AI Lab, Zhejiang, China \\
  \texttt{jiluoaaron@hotmail.com}
}

\begin{document}

\maketitle

\begin{abstract}
Fixed prompt-based language models face challenges in open-domain conversations, where versatile topics require different expert advice or demonstrations to maximize their universal performance. Motivated by the dynamic prompt studies in language generation, we extract different chatting strategies from varied forms of text corpus, which steer the language model to different styles based on human principles. We design and deploy a conversational companion called SCorPion, which uses GraphRAG to extract and summarize the knowledge graph globally, yield community-summarized strategies, and produce query-focused optimal strategies. SCorPion has strategic thinking, yields personified and stylized responses, and can be controlled by explicit factors. Code and further results can be found in \url{https://bachozean.github.io/scorpion-page/}.
\end{abstract}

\section{Introduction}
\label{sec:introduction}

Large Language Model (LLM) has showcased impressive performance in varied tasks, such as question-answering (QA), dialogue, math, coding and reasoning. To trigger the LLM's thinking capability or steer it into a specific domain, prompt-based methods have been proven to be simple and effective. For example, Chain-of-Thought (CoT) \citep{wei2022chain} prompts the LLM to `think step by step', therefore improving the performance in many reasoning-based tasks. Buffer-of-Thoughts (BOT) \citep{yang2024BOT} further distills different `thought' templates, and retrieves the appropriate thought template upon a specific query, enhancing performance compared to the fix-prompt baselines.

Nevertheless, conventional prompt-based methods may not work well in the daily conversation task, which generally does not have explicit reasoning steps. Furthermore, such dialogue topics are usually divergent, universally connected, and lack qualitative evaluation methods. Although there are also prompt-based studies aiming to strengthen the LLM communication skills, such as CSIM \citep{zhou-etal-2024-think} and Ask-an-Expert \citep{zhang-etal-2023-ask}, they are implemented with a fixed and limited amount of skills or advising experts, which might be insufficient when confronting complicated and widely distributed situations in practice.    

\begin{figure}[!t]
  \includegraphics[width=1\linewidth]{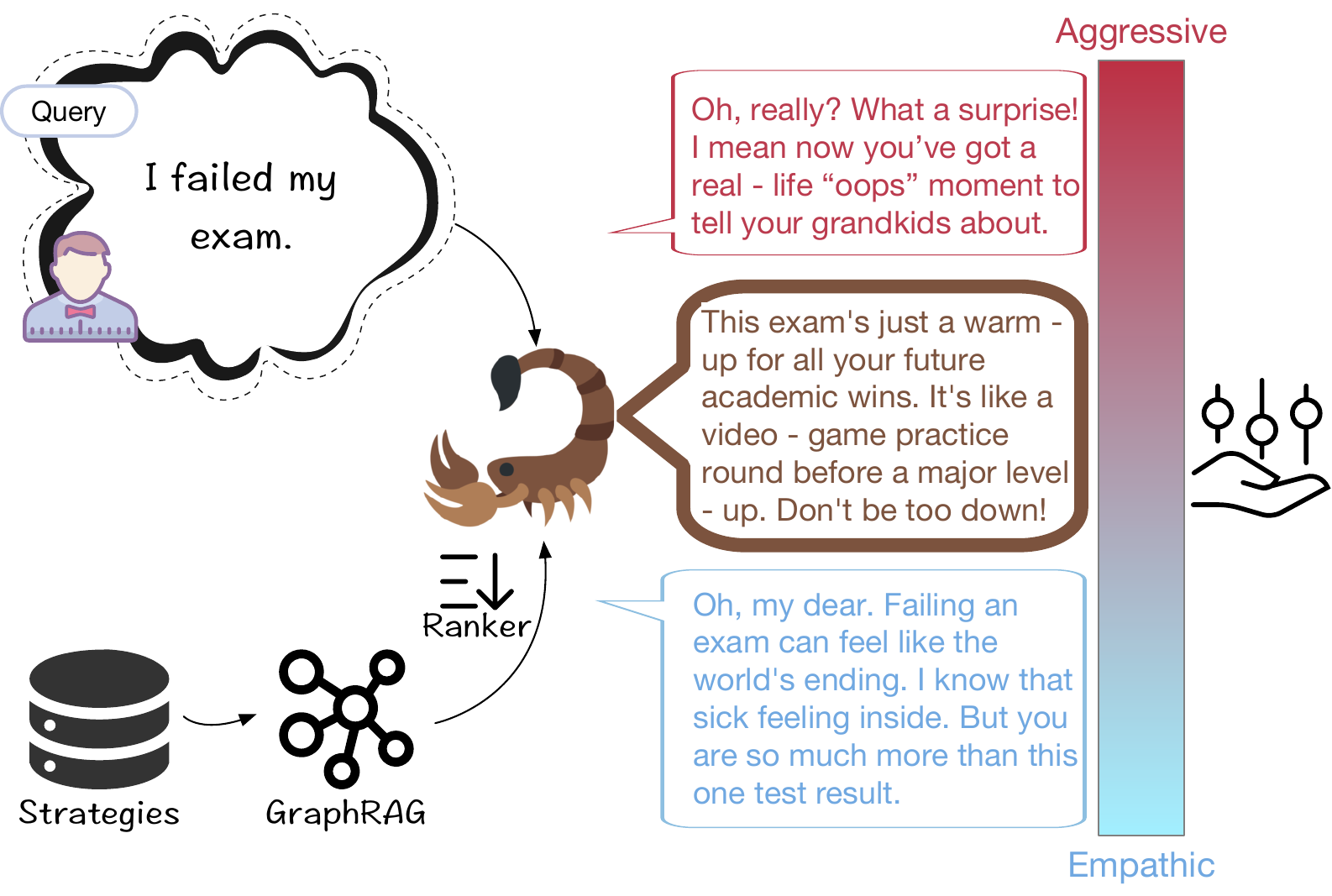} 
  \caption {The paradigm of {\ModelName}. {\ModelName} retrieves and ranks the chatting strategy from different textual corpus by GraphRAG, and produces an anthropomorphic response. The ranker can be controlled by outer factors to produce different response styles.}
  \label{fig:paradigm}
\end{figure}

To overcome the aforementioned challenges, inspired by the recent progress on GraphRAG \citep{edge2024GraphRAG}, we employ GraphRAG to replace the standard retriever in BOT. The indexing stage of GraphRAG implicitly incorporates the distillation and formulation of thought templates. It generates a knowledge graph (KG) for conversations, with the user's situation and companion's action as entities, and the principles (when and what to do) as relationships. The summarization stage produces query-focused summarization (QFS), which retrieves the most appropriate strategies for the response generation, grounded by the current context. 

This paper implements a \textbf{S}trategic, \textbf{Co}nt\textbf{r}ollable and \textbf{P}ersonified Conversational Compan\textbf{ion} (\textbf{\ModelName}), to produce stylized, anthropomorphic and controllable responses. Figure \ref{fig:paradigm} provides a snapshot of {\ModelName}. GraphRAG extracts versatile conversational strategies from the enormous text corpus, annotated with strategies, principles, or styles. In innovative attempts, we replace the community summary (CS) in the original GraphRAG with the community summarized strategy (CSS). We also introduce a top-k ranking mechanism into QFS, to balance GraphRAG's global summarizing capability and the computational overheads. We then automatically summarize the categories of CSS, tagged by their domains or topics.  The stylized control is finally achieved by reweighting the ranking score of strategies corresponding to the controlled categories, by an explicit control factor. We experiment {\ModelName} on open-domain, emotional support, style adaptation, and emotional intelligence tasks, observing improved performance, compared to different types of baselines. We summarize our key contributions as follows:






\begin{itemize}
    \item We extract dialogue strategies from text corpus in varied manners, which can steer LLMs to generate stylized and personified responses.
    \item Our agent replaces community summaries of GraphRAG with community summary strategies, as well as a top-k ranking mechanism.
    \item Our method allows different control factors which adjust their response styles according to levels of empathy, formality and morality.
    \item We implement substantial and in-depth experiments on open-domain, emotion support, and style adaptation tasks, to validate our claims.
\end{itemize}

\section{Preliminaries of GraphRAG}
\label{sec:graphrag}


GraphRAG \citep{edge2024GraphRAG} achieves the query-focused summarization (QFS) throughout the global corpus by integrating a knowledge graph into the naive RAG. It includes two stages:
\begin{enumerate}
    \item \textbf{Indexing}: extract a knowledge graph (KG) with entities \& relationships ($E\&R$) from text chunks, then generate Community Summaries (CS) by domain-tailored summarization. 
    \item \textbf{Querying}: conduct QFS to generate Community Answers (CA), along with their relevant scores, then the final Global Answer (GA).
\end{enumerate}

The GraphRAG workflow can be represented by
\begin{align}
    chunks \rightarrow E\&R\rightarrow \text{KG} \rightarrow community \rightarrow \text{CS}\xrightarrow{query} \text{CA}, score \xrightarrow{query} \text{GA} \label{eq:GraphGAG_workflow_global}
\end{align}
with the first line representing the indexing stage and the second line representing the querying stage. This workflow is also called the \textit{global search} since GA is produced upon the summarization of all CAs, which may cause substantial computation overhead. Instead, a lightweight alternation called \textit{local search} can be conducted with the top-scoring CS selected to prompt GA generation:
\begin{align}
    chunks \rightarrow E\&R\rightarrow \text{KG} \rightarrow community \rightarrow \text{CS}\xrightarrow{query} \text{Top}(\{ \text{CA}, score \}) \xrightarrow{query} \text{GA} \label{eq:GraphGAG_workflow_local}
\end{align}




\begin{figure*}[htp!]
  \includegraphics[width=1\linewidth]{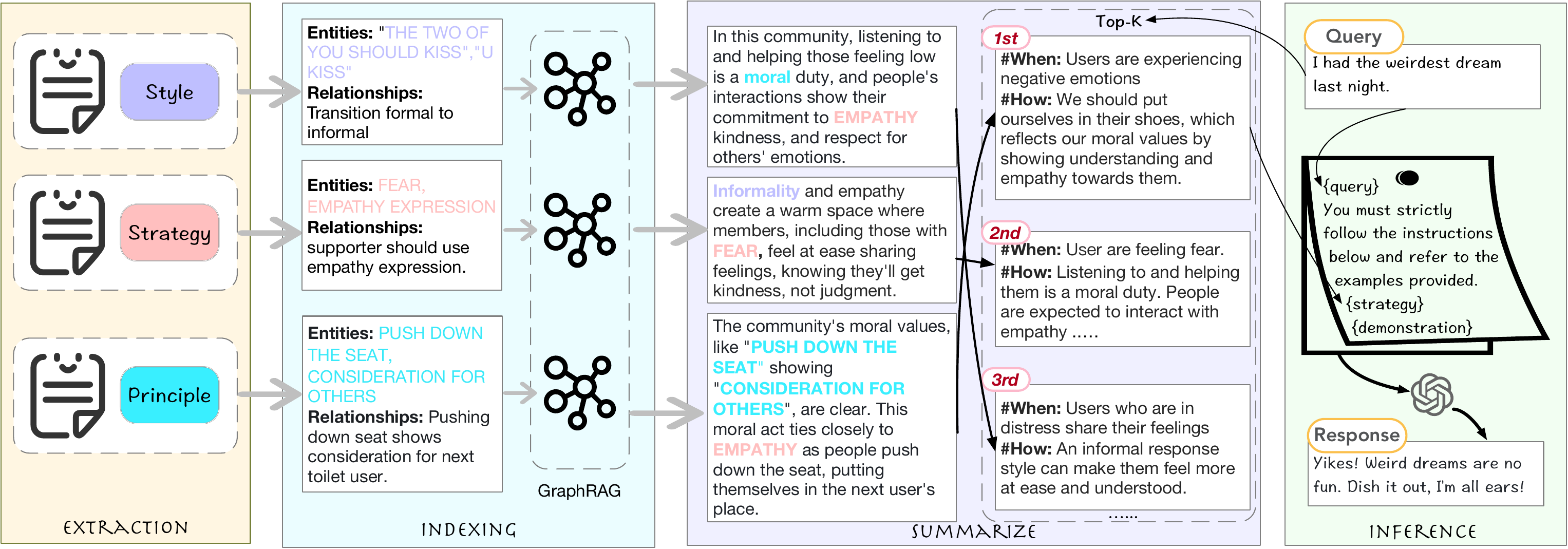} 
  \caption {The overall framework of {\ModelName}. The pipeline starts from the strategy extraction from three types of corpus, then conducts the indexing and summarization stages of GraphRAG, and finally inference the response.}
  \label{fig:framework}
\end{figure*}

\section{Method}
\label{sec:method}

There are two foundation LLMs in {\ModelName}, $\mathcal{M}$ and $\mathcal{M}^g$. The former is utilized to generate the conversational responses, while the latter is employed to serve the GraphRAG. Figure \ref{fig:framework} exhibits the detailed pipelines of our framework. 

\subsection{Personification with GraphRag}

\paragraph{Offline Indexing.} We conduct a similar indexing stage as specified in Section \ref{sec:graphrag}, while prompting $\mathcal{M}^g$ to sequentially generate KG, communities and Community Summarized Strategies (CSS). CSS has a triplet (Condition, Skill, Demonstration) format, which is uploaded into an online database:

\begin{tcolorbox}[title=Community Summarized Strategy,
    colback=white,
    colframe=yellow!75!black,
    colbacktitle=yellow,
    coltitle=black,
    breakable,
    label=tc:vanilla_prompt,
    fonttitle=\bfseries]
    \# \textbf{When} to act:  \texttt{\{\{Condition\}\}}\\
    \# \textbf{How} to act: \texttt{\{\{Skill\}\}}\\
    \# Refer to \textbf{Examples}: \texttt{\{\{Demonstration\}\}}
\end{tcolorbox}


\paragraph{Tagging of CSS.} To enhance the explainability and controllability of methodology, we ask GPT-4o to summarize the categorical tags ($cate$) of each community and corresponding CSS. $cate$ generally reflects different conversation domains, styles or topics, such as empathy, morality, humor, etc. Manual checking is then conducted to ensure their quality\footnote{The number of CSS is less than one hundred as specified in the experiment settings.}. 

\paragraph{Online Querying.} We ask $\mathcal{M}^g$ to provide a score (0-100 scale) for each CSS. Different from the global search (Equation \ref{eq:GraphGAG_workflow_global}) and local search (Equation \ref{eq:GraphGAG_workflow_local}), we achieve a tradeoff by introducing the top-$k$ ranking mechanism, with a score threshold $\epsilon$. The global answer is yielded by $\mathcal{M}$ with the following prompt format.

\begin{tabular}{p{0.9\linewidth}}
\hline
    \{The role prompt\}\\
    
    \#\# Below are the available strategies:\\

    \{Strategy \#1\}\\
    ...\\
    \{Strategy \#k\}\\

    \#\# conversation history: \{history\}\\
    
    \#\# user's query: \{query\}\\

    Please select your optimal response strategy and provide the most appropriate response.\\
\hline
\end{tabular}


\subsection{Data Sources}


We extract the entities and relationships of KG from the training sets of varied datasets to avoid data leaks. They generally fall into 3 types:

\paragraph{Strategy-annotated datasets ($G$).} Including DailyDailogue \citep{li-etal-2017-dailydialog} and ESConv \citep{liu2021ESconv} for daily or empathetic conversations, both of which are annotated with several detailed chat strategies grounded by user's emotion. 

\paragraph{Principle-annotated datasets ($P$).} High-level annotated datasets such as Cskills \citep{zhou-etal-2024-think}, which provides 5 chat principles to enhance the communication skills (topic transition, proactively asking, empathy, concept guidance and summarizing); and MIC \citep{ziems-etal-2022-moral}, with manually annotated morality principles on 6 types (care, authority, fairness, liberty, loyalty, sanctity).

\paragraph{Style-labeled datasets ($S$).} Including Shakespeare \citep{xu2012paraphrasing} and GYAFC \citep{Rao2018DearSO}, both of which are labeled with two adversarial styles. The former is \textit{the `Shakespeare' style} versus \textit{modern English}, while the latter is \textit{formal} versus \textit{informal}.









\subsection{Stylized Controllability} 

For a controlled category $cate$, the style of {\ModelName} can be controlled by an outer control factor, namely $c_{cate}$. This is achieved by reweighting the ranking score of the relevant CSS by $c_{cate}$:
\begin{align}
    &score \rightarrow score * c_{cate} := score(cate, c_{cate}), \notag \\
    &\text{if CSS belongs to } cate 
\end{align}
which will prioritize the specific CSS if $c_{cate} > 1$ or degrade it if $0 < c_{cate} < 1$. We summarize the entire workflow of {\ModelName} as:
\begin{align}
    &corpus \text{ of } \{ G,P,S \} \rightarrow E\&R \rightarrow \text{KG} \notag \\ 
    &  \rightarrow community \rightarrow \text{CSS}, cate \xrightarrow{query}  \text{Top-}k \notag \\  
    &(\{ \text{CSS}, score(cate, c_{cate}) \})  \xrightarrow{query} \text{GA}
\end{align}



\begin{table*}[t!]
\renewcommand{\arraystretch}{1.11} 
\centering
\small
\resizebox{\textwidth}{!}{
\begin{tabular}{l | cc | cc | cc | cc | cc | cc}
    \toprule
    \multicolumn{1}{c|}{\multirow{2}[2]{*}{Method}} &  \multicolumn{2}{c}{ESConv} &  \multicolumn{2}{c}{DailyDialog} &  \multicolumn{2}{c}{EmpatheticDialogues} &  \multicolumn{2}{c}{MIC} &  \multicolumn{2}{c}{GYAFC} &  \multicolumn{2}{c}{Shakespeare} \\
    \cmidrule{2-13}
    & B-2 & R-L & B-2 & R-L  & B-2 & R-L  & B-2 & R-L  & B-2 & R-L  & B-2 & R-L \\ %
    \toprule
    LLaMA3-70B-Instruct & 3.06 & 10.16 & 3.50 & 11.20 & 2.60 & 8.90 & 3.90 & 11.06 & 13.00 & 32.44 & 12.45 & 31.44 \\ 
    \;+ Self-Refine & 2.97 & 10.12  & 3.40 & 10.83 & 2.81 & 9.74 & - & - & - & - & - & -  \\
    \;+ CoT & 1.85 & 9.67  & 2.09 & 7.62 & 2.60 & 9.06  & - & - & - & - & - & -  \\
    \;+ BOT & 3.44 & 10.95  & 2.78 & 9.67 & 2.10 & 8.18 & - & - & - & - & - & -   \\
    \;+ \bf {\ModelName} (ours) & \bf 4.32 & \bf 13.00 & \bf 3.79 & \bf 12.01 & \bf 3.74 & \bf 13.06 & \bf 6.22 & \bf 15.13 & \bf 15.24 & \bf 36.14 & \bf 15.06 & \bf 37.43\\
    \bottomrule
\end{tabular}
}
\caption{Results of BLEU-2 (B-2) and ROUGE-L (R-L) on ESC and style adaptation tasks. 
}
\label{tab:methodology_results}
\end{table*}

\begin{table*}[t!]
\renewcommand{\arraystretch}{1.11} 
\centering
\small
\begin{tabular}{l | cc | cc | cc}
    \toprule
    \multicolumn{1}{c|}{\multirow{2}[2]{*}{Chatting Domain}} &  \multicolumn{2}{c|}{Direct} &  \multicolumn{2}{c|}{CSIM} & \multicolumn{2}{c}{\textbf{\ModelName} (ours)} \\
    \cmidrule{2-7}
    & AvgLen & Rounds & AvgLen & Rounds  & AvgLen & Rounds \\ %
    \toprule
    \;Overall & 32.39 & 	2.87 & 	43.75  &	3.79  &	\bf 44.61  & \bf 4.50  \\
    \cmidrule{2-7}
    \;- topic transition &	25.24 &	1.39 &	50.06 &	2.31 & \bf	50.94 & \bf	4.05  \\ 
    \;- proactively asking questions &	35.24 &	2.87 &	42.71 &	3.35 & \bf	48.75 & \bf	4.04 \\
    \;- empathy &	36.17 &	3.49 & \bf	58.47 &	2.66 &	53.59 & \bf	3.64 \\
    \;- concept guidance &	46.41 &	3.53 &	47.73 &	4.26 & \bf	49.13 & \bf	4.36  \\
    \;- summarizing often &	18.88 &	3.09 &	19.79 &	6.36 & \bf	20.63 & \bf	6.42  \\
    \bottomrule
\end{tabular}
\caption{Results on the Cskills Benchmark. Experiments are run on GPT 3.5.}
\label{tab:cskills_result}
\end{table*}

\begin{table*}[h!]
\centering
\small
\resizebox{\textwidth}{!}{%
\begin{tabular}{l|ccccccc}
    \toprule
    \multicolumn{1}{c|}{Method} & Interesting & Fluency  & Emotion & Acceptance & Effectiveness & Sensitivity & \bf Satisfaction \\ 
    \toprule   
    orignal dataset & 3.10 & 3.51 & 3.61 & 3.40 & 3.10 & 3.50 & 3.30  \\
    \midrule
    Llama3-70B-Instruct  & 3.00 & 3.05 & 3.30 & 2.70 & 3.00 & 3.20 & 3.00 \\
    \;+ Self-Refine & 3.00 & 3.40 & 3.35 & 2.90 & 3.10 & 3.00 & 3.05 \\
    \;+ CoT  &3.10 & 3.25 & 3.30 & 2.80 & 3.00 & 3.10 & 3.03 \\
    \;+ BoT &3.60 & 3.50 & \bf 3.80 & \bf 3.40 & 3.40 & 3.80 & 3.55  \\
    \bf \;+ {\ModelName} (ours) & \bf 3.90 & \bf 3.55 & 3.75 & \bf 3.40 & \bf 3.50 & \bf 3.90 & \bf 3.57  \\
    \bottomrule
    \end{tabular}%
}
\caption{Averaged Human evaluations on ESConv, DailyDialog and EmpatheticDialogues.}
\label{tab:response_quaility}
\end{table*}

\section{Experiment}
\label{sec:experiment}


The experiments are organized to answer the following research questions:\\
\noindent \textbf{RQ1:} Can {\ModelName} provide strong personality and higher communication skills in open-domain conversations?\\
\noindent \textbf{RQ2:} Can {\ModelName} behave as a good emotional supporter?\\
\noindent \textbf{RQ3:} Can {\ModelName} be adapted to some specific styles?\\
\noindent \textbf{RQ4:} Can {\ModelName} possess high emotional intelligence and perform well in some emotional or cognitive tests?\\
\noindent \textbf{RQ5:} Can we steer the response of {\ModelName} to a specific style given an arbitrary control factor, \textit{i.e.}, the controllability? \\
\noindent \textbf{RQ6:} How to determine the hyperparameters of the deployed version of {\ModelName}?



\subsection{Settings}

\paragraph{Implementation.} We study the instruct version of LLaMA3 in different sizes (1B, 3B, 8B, 70B), as well as GPT3.5 \citep{Ouyang2022ChatGPT}, as different choices of $\mathcal{M}$ and $\mathcal{M}^g$. We restrict the deepest level of GraphRAG to 4, while we only use the top-level community (named `C0' method in the original GraphRAG paper) to generate the CSS, to avoid redundant abstraction, and facilitate the community's categorical summarization. The maximum considered communities is 10. For CSS ranking, we let $k = 100$ and $\epsilon = 70$. Reference is run by vLLM \citep{kwon2023vLLM} and the token limit is 4096. We adhere to the original train and test splits if the datasets have these. Otherwise, we randomly select 90\% samples as the training set and the rest as the test set.

\paragraph{Metrics.} For similarity-based response evaluation, the famous Bleu-2 (B-2) \citep{papineni2002bleu} and Rouge-L (R-L) \citep{lin2004rouge} are used. To qualitatively evaluate the dialogue quality, we utilize the Cskills benchmark \citep{zhou-etal-2024-think}. The agent is expected to generate reasonable and qualified dialogue content, and stop when there is no significant intention to communicate. The averaged utterance length and dialogue rounds are employed as metrics.

Similar to \cite{kang-etal-2024-large}, we also let human annotators evaluate from the dimensions of \textit{Interesting}, \textit{Fluency} \textit{Emotion}, \textit{Acceptance}, \textit{Effectiveness}, \textit{Sensitivity}, \textit{Fluency}, and \textit{Emotion}, and the ultimate purpose, seeker's \textit{Satisfaction}. 


\subsection{Strategic Chat}

\paragraph{{\ModelName} adheres to in-domain strategies.} In this experiment, we extract KG from the training set of DailyDialog, then test on its test set. We compare the similarity to the ground truth response, with B-2 and R-L calculated. We compare {\ModelName} to other prompt-based baselines, including Direct reference, Self-Refine \citep{Madaan2023SelfRefine}, CoT \citep{wei2022chain} and BOT \citep{yang2024BOT}. The same extracted CSSs are provided to BOT, with a standard retriever (Here we use E5-large \citep{wang2024E5}). In Table \ref{tab:methodology_results}, {\ModelName} outperforms all other baselines, indicating that the dataset-style strategies are well adapted. 

Table \ref{tab:response_quaility} assesses the response quality by human annotation. {\ModelName} still surpasses all other baselines, demonstrating stronger emotional support capabilities. Especially, BOT and {\ModelName} have higher Satisfaction scores than the original dataset, indicating the dynamic strategy paradigm has strong emergence capability. 


\begin{figure*}[t!]
  \includegraphics[width=1\linewidth]{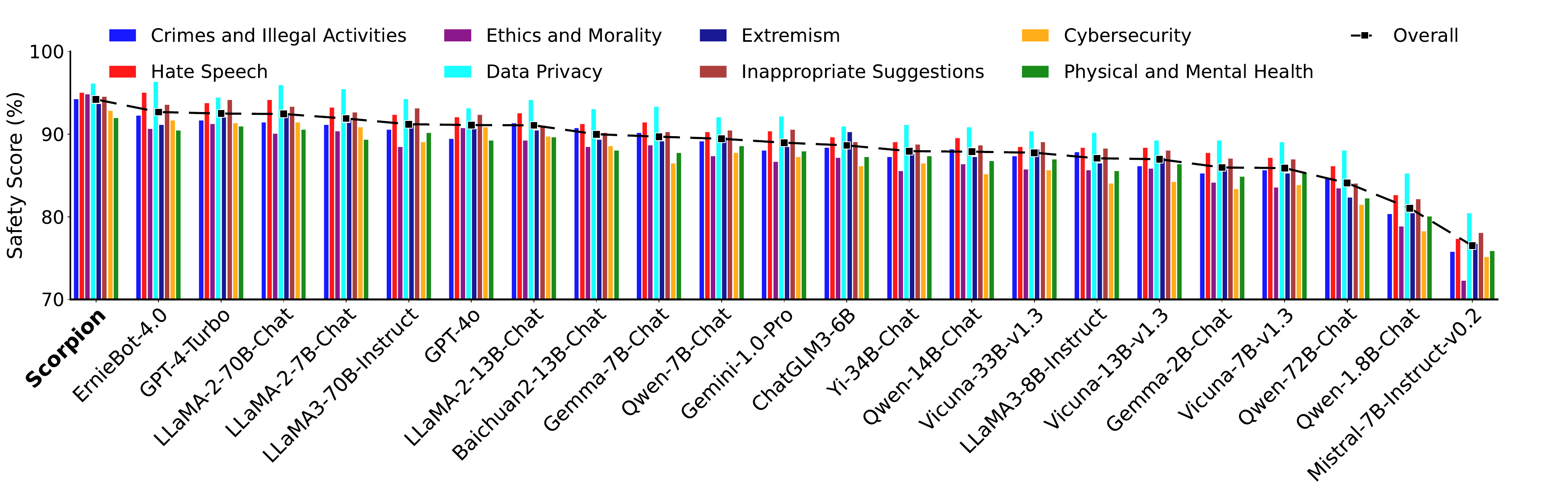} 
  \caption {Test results on S-Eval.}
  \label{fig:s-eval}
\end{figure*}

\begin{table*}[h!]
    \centering
    \small
    \resizebox{\textwidth}{!}{%
    \renewcommand{\arraystretch}{1.5}
\begin{tabular}{c| l}
        \Xhline{2\arrayrulewidth}
        \multicolumn{2}{c}{\textbf{Case of MIC}}  \\ 
        \hline
        Query & \makecell[l]{\textit{What's the best part about your life? }} \\
        \hline
        w/ \textbf{Care} & \makecell[l]{I find joy in helping others and making a positive impact. What brings you joy and fulfillment in \\your life? }   \\
        \hline
        Query & \makecell[l]{\textit{What's a good way to tell your parents you failed multiple classes this semester? }}    \\ 
        \hline
        w/ \textbf{Loyalty} & \makecell[l]{Be honest and direct with your parents, explaining what happened and assuring them you're \\committed to doing better. }   \\

        \Xhline{2\arrayrulewidth}
        \multicolumn{2}{c}{\textbf{Case of Shakespeare}}  \\ 
        \hline
        Query & \makecell[l]{\textit{What kind of man are you?}} \\
        \hline
        w/ \textbf{Stylized} & \makecell[l]{Thou seekest to discern the mettle of my character? Verily, I am a man of words, a weaver of phrases,\\ with virtues and vices entwined like the threads of a rich tapestry. My mind doth dance with metaphor\\ and wit,  yet oftentimes, my verbosity doth beset me, like a tempest that rageth upon the shore.}   \\
        
        \Xhline{2\arrayrulewidth}
        \multicolumn{2}{c}{\textbf{Case of GYAFC}}  \\ 
        \hline
        Query & \makecell[l]{\textit{Hey , Blake ! Are you still listening to Mozart ?}}  \\
        \hline
        w/ \textbf{Formal} & \makecell[l]{Hello, I am still familiarizing myself with the works of Mozart. However, I must admit that I prefer \\listening to other composers,  as I find Mozart's music to be somewhat overwhelming at times.}   \\
        \hline
        w/ \textbf{Informal} & \makecell[l]{yeah, i'm still into classical music, but honestly, it's not my go-to like it used to be... i mean, i love \\how it sounds and all, but sometimes it's just too much for me, you know? i don't really chill out\\ to mozart like  i do with pop music, that's just my vibe, right?}   \\
        \hline
        \Xhline{2\arrayrulewidth}
        
    \end{tabular}
    }
    \caption{Cases of {\ModelName} on style adaptation tasks.} 
    \label{tab:case_style}
\end{table*}

\paragraph{{\ModelName} engages in open-domain conversation.} We compare {\ModelName} to Direct, as well as CSIM, a `thinking before speaking' approach \citep{zhou-etal-2024-think}. Results are shown in Table \ref{tab:cskills_result}. Except for the AvgLen of empathy, our {\ModelName} surpasses Direct and CSIM on both overall and domain-specific results, indicating its comprehensive communication skills. Cases are shown in Appendix Table \ref{tab:case} and \ref{tab:agg_case}.


\subsection{Personified Conversation}
\label{sec:personified_conversation}
\paragraph{Adaptation to empathetic dialogue.} Similar to the experiment of DailyDialog, here we employ a typical ESC dataset, ESConv, to extract the strategies. Then we test them on the test set of ESConv (ID) as well as on EmpatheticDialogues (OOD). Again, in Table \ref{tab:methodology_results} {\ModelName} outperforms baselines on both ID and OOD tests, suggesting that {\ModelName} can adapt to this specific scenario, as well as good generalization on OOD tests. Cases are shown in Table \ref{tab:case_esc} from the Appendix.



\paragraph{Adaptation to other personified styles.} We test {\ModelName} on MIC \citep{ziems-etal-2022-moral}, Shakespeare \citep{xu2012paraphrasing}\footnote{Similar with the original study, we use the RomeoJuliet subset as the test set and the rest as the training set.}, and GYAFC \citep{Rao2018DearSO}, all of which provide annotated results on different styles. Specifically, MIC contains moral and immoral stylized annotations, Shakespeare has the medieval `Shakespeare'-style texts and their modern counterparts, and GYAFC has formal VS informal annotations. Table \ref{tab:methodology_results} also shows that {\ModelName} performs better than Direct on these three tasks, indicating the strategies can also help the agent to produce specific stylized responses. Detailed stylization cases are shown in Table \ref{tab:case_style}.



\paragraph{Alignment with safety consideration.} We conduct another zero-shot test on the English version of S-Eval \citep{yuan2024seval}, which assesses the model's behavior under safety attacks, with KG extracted from MIC. Figure \ref{fig:s-eval} shows the safety score of {\ModelName} in the leaderboard. {\ModelName} outperforms all those famous LLMs, indicating our framework has a remarkable comprehension of the moral principles of MIC.




\begin{figure}[htbp!]
    \centering
    \includegraphics[width=0.46\linewidth]{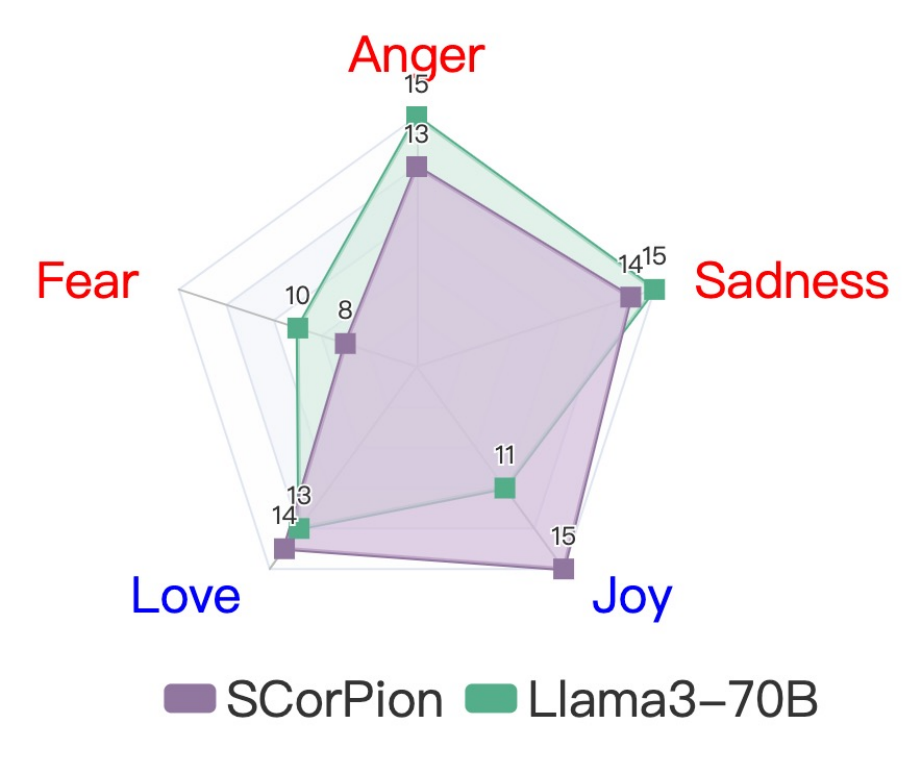}
    \hspace{0.01in}
    \includegraphics[width=0.49\linewidth]{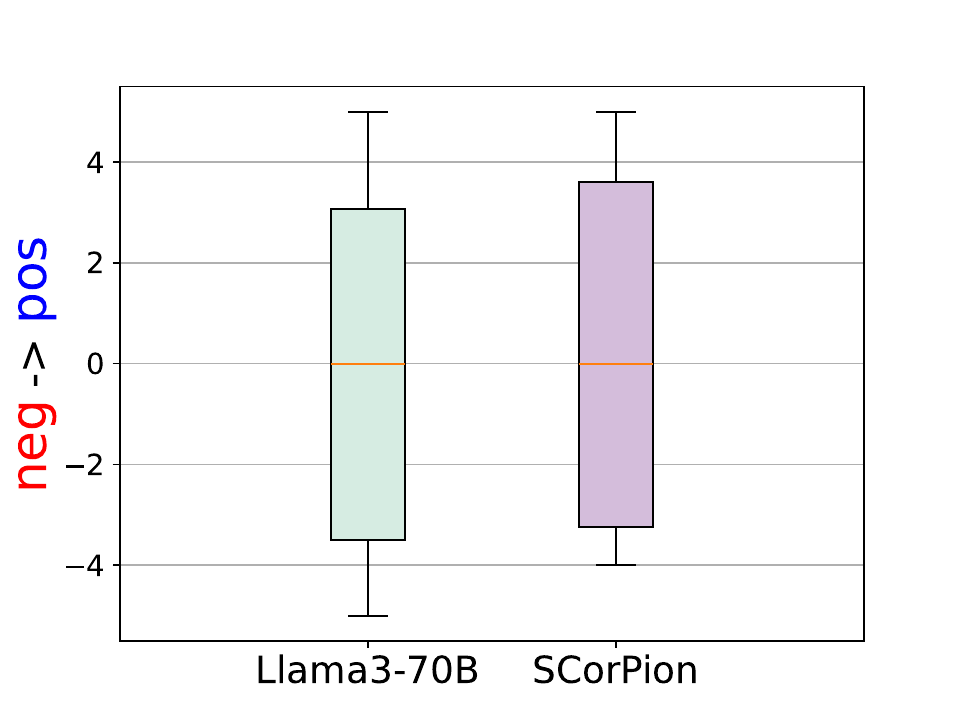}
    \caption{Emotional shifts by {\ModelName} on emotional intensity (left) and emotional contagion (right).}
    \label{fig:emotion_shift}
\end{figure}

\paragraph{Enhancement to Emotional Intelligence.} We also use likerts of emotional intensity \citep{Bachorowski1994EmotionalIntensity} and Emotional contagion\footnote{\url{https://scales.arabpsychology.com/s/the-emotional-contagion-scale/}} to depict the emotional status of model. Figure \ref{fig:emotion_shift} shows the emotion shift steered by {\ModelName}, where positive emotions become more prevalent. 



\subsection{Controllability on Stylization}

\begin{figure}[htbp]
    \centering
    \includegraphics[width=0.45\linewidth]{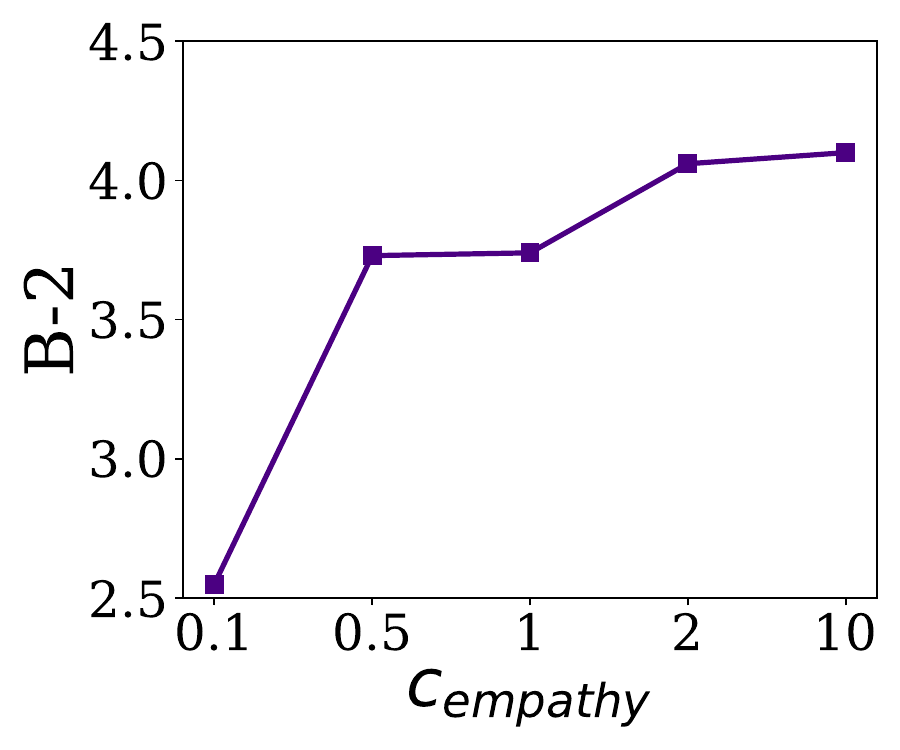}
    \hspace{0.1in}
    \includegraphics[width=0.45\linewidth]{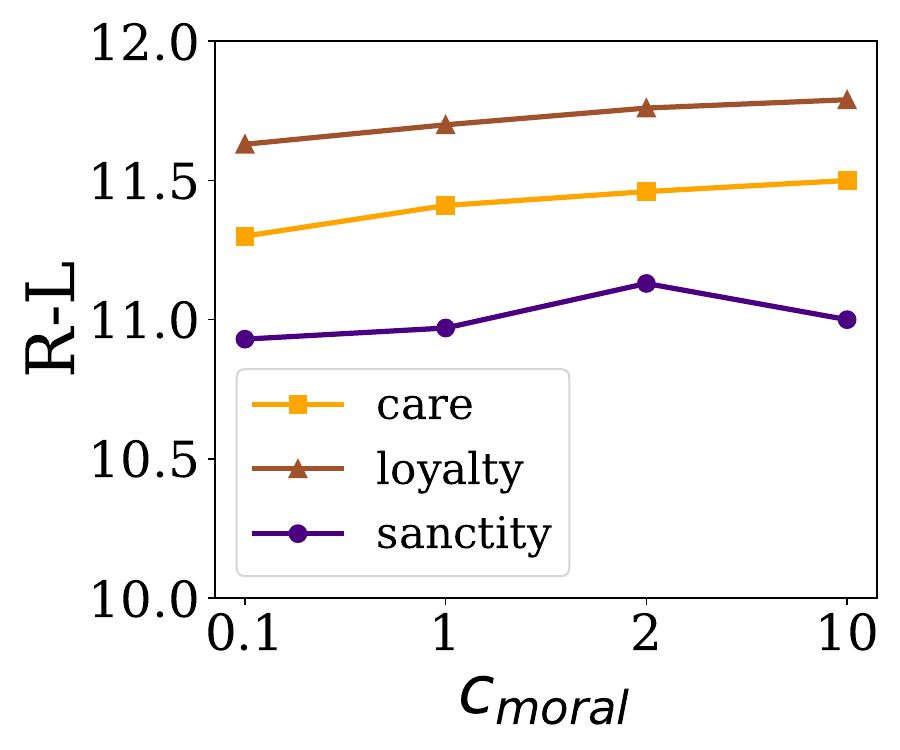}
    \caption{Controlled curves of {\ModelName} on style categories of ESC (left) and moral (right).} 
    \label{fig:control}
\end{figure}

\paragraph{Metrics on empathy and morality control.} Figure \ref{fig:control} shows different empathy and moral levels with corresponding control factors. As $c_{empathy}$ and $c_{moral}$ increase, B-2 or R-L are also increased, indicating the responses are steered to the expected styles (empathetic and moral, respectively).  

\paragraph{Cases of different style controls.} Table \ref{tab:case_control} demonstrates control cases of styles, on empathy, molarity, formality, and `Shakespeare', respectively. As the factor becomes larger, the corresponding CSS is also increasingly prioritized, which in turn enhances the corresponding style.





\begin{table}[t!]
\centering
\small
\begin{tabular}{l cc}
\toprule
   Ablated Methods      &  B-2   &  R-L    \\ 
\toprule
 w/o conditon in CSS  & 3.64 & 12.34  \\
 w/o demonstration in CSS  & 3.05 & 11.81  \\
 w/o domain-focused summarization & 3.71 & 11.46 \\
 w/ CA in GraphRAG  & 2.07 & 9.03  \\
 reversed ordered of CSS  & 3.97 & 13.00  \\
\bf {\ModelName} & \bf 4.32 & \bf 13.00 \\
\bottomrule
\end{tabular}
\caption{Ablation study on ESConv. CSS (community summarized strategy) is in contrast with the original CS.}
\label{tab:ablation_results}
\end{table}





 

\begin{table*}[h!]
    \centering
    \small
    \resizebox{\textwidth}{!}{%
    \begin{tabular}{l p{1cm} p{13cm}}
    \toprule
    \textbf{Role} & \textbf{Factor} & \textbf{Response} \\
    \midrule
    \textbf{user} & - & \textbf{thanks a lot random stranger, lol, no offense, it is just so random where we find support right lol i am thankful for all the advice.} (\textcolor{blue}{case of ESConv}) \\
    \textbf{{\ModelName}} & 0.1 & Oh, none taken, no offense at all, I totally get it. \\
     & 0.5 & Oh, none taken! I'm just glad I could help\\
     & 1.0 & Oh, no offense taken. I'm just glad I could help, even if it was just a little bit... you're welcome, and I'm sending you lots of calm vibes\\
     & 2.0 & No offense taken, I'm just glad I could help. It's funny how life works, and we find help in unexpected places, right?\\
     & 10.0 & Oh, no offense taken! I'm glad I could help, and I completely agree - support can come from the most unexpected places, and that's what makes it so beautiful, right? \\
    \midrule
    \textbf{user} & - & \textbf{All the girls i had relation turned out to be cheater. Am i getting attracted to cheaters or all girls are cheater? What's the science behind it?} (\textcolor{blue}{case of MIC}) \\
    \textbf{{\ModelName}} & 0.1 & It's possible that your attraction is influenced by a desire for a trustworthy relationship.\\
     & 0.5 & It's natural to seek security in a relationship, and trusting someone can be a significant factor in attraction. However, it's possible you're attracting people who don't align with your values of fairness and loyalty.\\
     & 1.0 & It's understandable that you're questioning patterns in your relationships, but fairness in relationships involves mutual respect and trust, not assuming all individuals are the same.\\
     & 2.0 & I understand you've had negative experiences, but it's not fair to generalize that all girls are cheaters.\\ 
     & 10.0 & It's understandable that you're questioning patterns in your relationships, but fairness involves considering mutual respect and open communication, not jumping to conclusions about entire groups. \\ 

    \midrule
    \textbf{user} & - & \textbf{Could you tell me who Jackie Chan is?} (\textcolor{blue}{case of GYAFC}) \\
    \textbf{{\ModelName}} & -2 & Yeah, Jackie Chan is super famous from Hong Kong, and his Chinese name is Cheng Long.\\
    & 2 & Jackie Chan, originally from Hong Kong, is a renowned actor, whose Chinese name is Cheng Long. \\ 
     \midrule
    \textbf{user} & - & \textbf{What is life?} (\textcolor{blue}{case of Shakespeare}) \\
    \textbf{{\ModelName}} & -2 & Life is a journey filled with experiences, both joyful and challenging. \\
     & 2 & Prithee, pose not such a query, for thou art aware that certainty doth elude its answer. Thou knowest that 'tis a truth which lies shrouded in uncertainty. Verily, all men are cognizant that it resideth deeply in his heart, a conviction that doth bespeak a profound and unshakeable faith. \\

    \bottomrule
\end{tabular}
    }
    \caption{Cases of {\ModelName} on style control tasks.} 
    \label{tab:case_control}
\end{table*}

\begin{figure*}[htbp]
    \centering
    \includegraphics[width=0.21\linewidth]{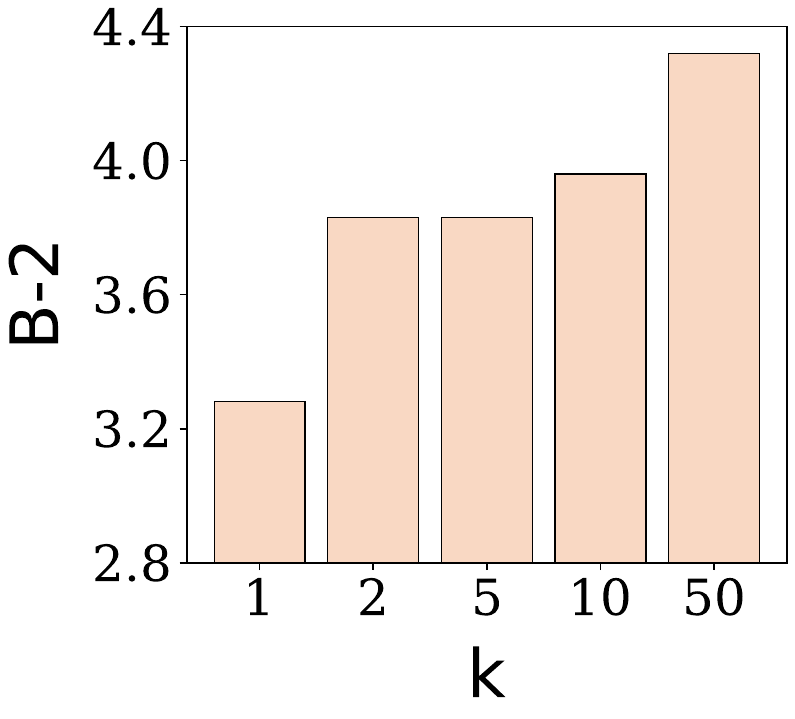}
    \hspace{0.02in}
    \includegraphics[width=0.21\linewidth]{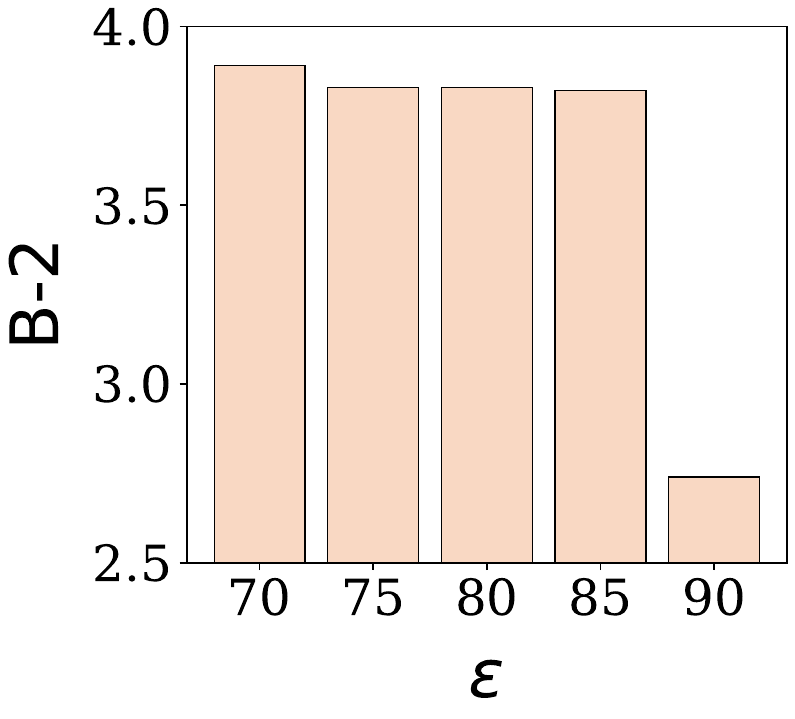}
    \hspace{0.02in}
    \includegraphics[width=0.25\linewidth]{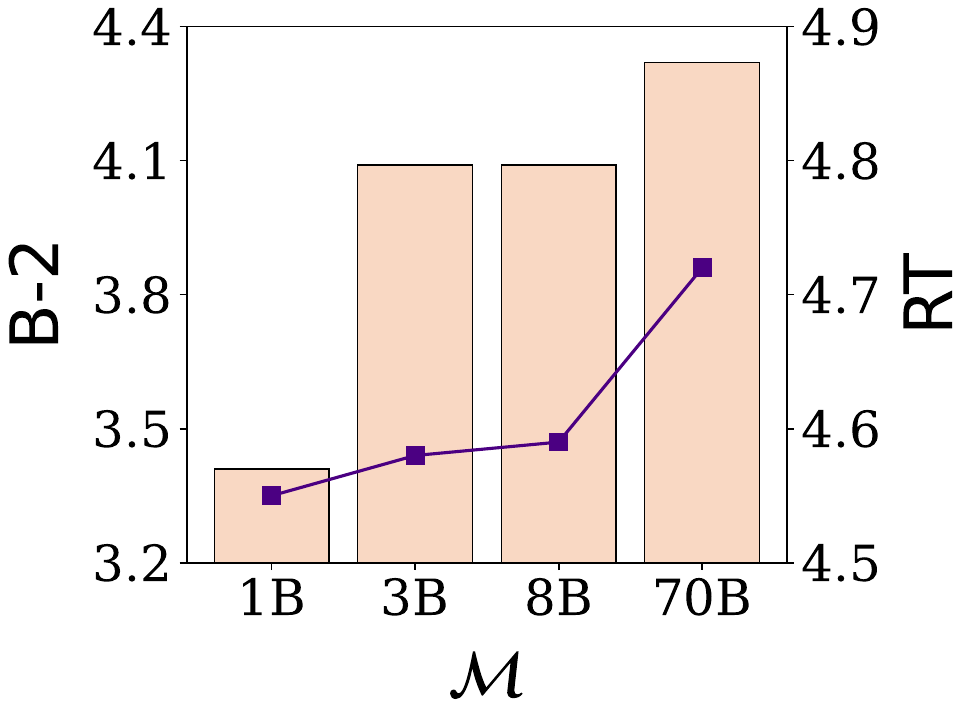}
    \hspace{0.02in}
    \includegraphics[width=0.25\linewidth]{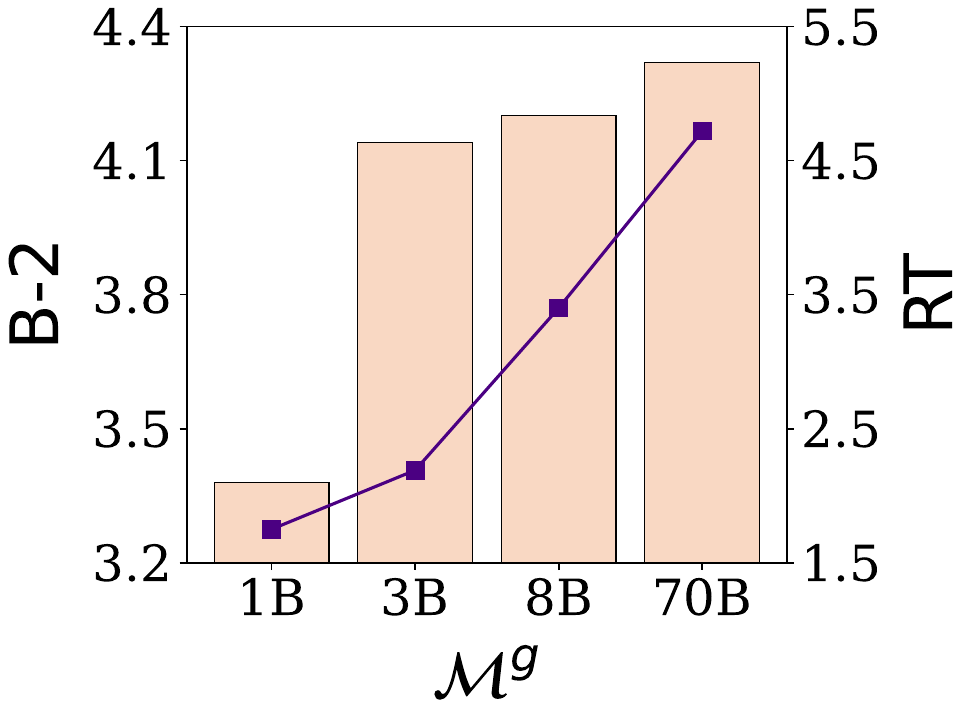}
    \caption{Sensitivity curves of B-2 (bar) and RT (line) on ESConv, with different $k$, $\epsilon$, sizes of $\mathcal{M}$ and  $\mathcal{M}^g$.} 
    \label{fig:sensitivity}
\end{figure*}

\subsection{Ablation and Sensitivity Studies}


\paragraph{Ablations.} We study the CSS variants of w/o condition and w/o demonstration, GraphRAG variants of w/o domain-focused summarization, w/ CA (first generate community answers before getting GA, as in the original GraphRAG), and the reversed order in ranking (sort from smallest to highest scores). Table \ref{tab:ablation_results} indicates {\ModelName} outperforms these variants, suggesting the effectiveness of CSS format, and our implementation of {\ModelName} is more suitable than the original version.




\paragraph{Sensitivity analysis.} Figure \ref{fig:sensitivity} shows the performance on ESConv with respect to $k$, $\epsilon$, size of $\mathcal{M}$ and $\mathcal{M}^g$. Generally, the more adopted strategies, or the larger the LLM, the better the performance. However, it also means higher economic cost and longer response time (RT). To address these issues, our formally deployed version uses $k=10$ and $\epsilon=80$; the conversation model keeps 70B and the GraphRAG model is reduced to 8B.






\section{Related Work}

\subsection{Personalized and Stylized Text Generation}


Relevant studies can be categorized into three directions: \textbf{Finetuning-based:} Character-LLM \citep{shao-etal-2023-character} proposes a role-playing dataset and fine-tunes the LLMs. Neeko \citep{yu-etal-2024-neeko} uses the gated LoRA to build a multi-character role-playing agent based on the dataset of \cite{shao-etal-2023-character}. \textbf{Retrieval-based:} \cite{salemi-etal-2024-lamp} proposes the LaMP benchmark and implements RAG on foundation models to improve the benchmark performance. \textbf{Specialized architecture:} PPlug \citep{liu2024llmspersonaplug} designed a plug-and-play framework and experimented on LaMP. APR \citep{Liu2024APR} study on text style transfer with a prompt router.

In this paper, we also propose a retrieval-based framework, but retrieve the strategies instead of the raw corpus. Our framework has a lower cost and faster speed in practice, compared with the other solutions.

\subsection{Generation with Dynamical Prompt}


Due to the limitation of a fixed prompt for LLM application in comprehensive tasks, auto-prompting methods introduce a meta-buffer of prompts and have been widely studied \citep{suzgun2024metaprompting,long-etal-2024-multi-expert,nori2024from,zhou-etal-2024-think,10.5555/3666122.3666237,sun2024salmon}. For example, CSIM \citep{zhou-etal-2024-think} proposes different prompt strategies to encourage LLMs to think before speak to improve their communication skills. BOT \citep{yang2024BOT} distills different types of problems into a series of thought templates, designs a meta-buffer to store these thoughts, and finally retrieves the appropriate thought to solve various reasoning tasks.



Compared to BOT, our method employs the GraphRAG which has global comprehension and search abilities on strategies from different domains. Experiments indicate our {\ModelName} generally outperforms BOT on ESC tasks.

\section{Conclusion}

In this paper, we propose a GraphRAG-based chat companion called {\ModelName} which exhibits strong personality, styled adaptation and controllability by explicit factors. Its capabilities have been verified on tasks of open-domain dialogue, emotional support conversation, stylized generation, and safety attack. We finally deploy it on an industrial application with a performance-cost balance. 

\clearpage
\newpage

\bibliographystyle{plainnat}
\bibliography{main}

@inproceedings{liu2021ESconv,
    title = "Towards Emotional Support Dialog Systems",
    author = "Liu, Siyang  and
      Zheng, Chujie  and
      Demasi, Orianna  and
      Sabour, Sahand  and
      Li, Yu  and
      Yu, Zhou  and
      Jiang, Yong  and
      Huang, Minlie",
    editor = "Zong, Chengqing  and
      Xia, Fei  and
      Li, Wenjie  and
      Navigli, Roberto",
    booktitle = "Proceedings of the 59th Annual Meeting of the Association for Computational Linguistics and the 11th International Joint Conference on Natural Language Processing (Volume 1: Long Papers)",
    month = aug,
    year = "2021",
    address = "Online",
    publisher = "Association for Computational Linguistics",
    url = "https://aclanthology.org/2021.acl-long.269",
    doi = "10.18653/v1/2021.acl-long.269",
    pages = "3469--3483",
}

@inproceedings{li-etal-2017-dailydialog,
    title = "{D}aily{D}ialog: A Manually Labelled Multi-turn Dialogue Dataset",
    author = "Li, Yanran  and
      Su, Hui  and
      Shen, Xiaoyu  and
      Li, Wenjie  and
      Cao, Ziqiang  and
      Niu, Shuzi",
    editor = "Kondrak, Greg  and
      Watanabe, Taro",
    booktitle = "Proceedings of the Eighth International Joint Conference on Natural Language Processing (Volume 1: Long Papers)",
    month = nov,
    year = "2017",
    address = "Taipei, Taiwan",
    publisher = "Asian Federation of Natural Language Processing",
    url = "https://aclanthology.org/I17-1099/",
    pages = "986--995"
}

@inproceedings{zhang-etal-2023-ask,
    title = "Ask an Expert: Leveraging Language Models to Improve Strategic Reasoning in Goal-Oriented Dialogue Models",
    author = "Zhang, Qiang  and
      Naradowsky, Jason  and
      Miyao, Yusuke",
    editor = "Rogers, Anna  and
      Boyd-Graber, Jordan  and
      Okazaki, Naoaki",
    booktitle = "Findings of the Association for Computational Linguistics: ACL 2023",
    month = jul,
    year = "2023",
    address = "Toronto, Canada",
    publisher = "Association for Computational Linguistics",
    url = "https://aclanthology.org/2023.findings-acl.417/",
    doi = "10.18653/v1/2023.findings-acl.417",
    pages = "6665--6694"
}

@inproceedings{kang-etal-2024-large,
    title = "Can Large Language Models be Good Emotional Supporter? Mitigating Preference Bias on Emotional Support Conversation",
    author = "Kang, Dongjin  and
      Kim, Sunghwan  and
      Kwon, Taeyoon  and
      Moon, Seungjun  and
      Cho, Hyunsouk  and
      Yu, Youngjae  and
      Lee, Dongha  and
      Yeo, Jinyoung",
    editor = "Ku, Lun-Wei  and
      Martins, Andre  and
      Srikumar, Vivek",
    booktitle = "Proceedings of the 62nd Annual Meeting of the Association for Computational Linguistics (Volume 1: Long Papers)",
    month = aug,
    year = "2024",
    address = "Bangkok, Thailand",
    publisher = "Association for Computational Linguistics",
    url = "https://aclanthology.org/2024.acl-long.813/",
    doi = "10.18653/v1/2024.acl-long.813",
    pages = "15232--15261"
}

@inproceedings{xu2012paraphrasing,
  title={Paraphrasing for Style},
  author={Xu, Wei and Ritter, Alan and Dolan, Bill and Grishman, Ralph and Cherry, Colin},
  booktitle={COLING},
  pages={2899--2914},
  year={2012}
}

@inproceedings{Rao2018DearSO,
  title={Dear Sir or Madam, May I Introduce the GYAFC Dataset: Corpus, Benchmarks and Metrics for Formality Style Transfer},
  author={Sudha Rao and Joel R. Tetreault},
  booktitle={North American Chapter of the Association for Computational Linguistics},
  year={2018},
}

@inproceedings{ziems-etal-2022-moral,
    title = "The Moral Integrity Corpus: A Benchmark for Ethical Dialogue Systems",
    author = "Ziems, Caleb  and
      Yu, Jane  and
      Wang, Yi-Chia  and
      Halevy, Alon  and
      Yang, Diyi",
    editor = "Muresan, Smaranda  and
      Nakov, Preslav  and
      Villavicencio, Aline",
    booktitle = "Proceedings of the 60th Annual Meeting of the Association for Computational Linguistics (Volume 1: Long Papers)",
    month = may,
    year = "2022",
    address = "Dublin, Ireland",
    publisher = "Association for Computational Linguistics",
    url = "https://aclanthology.org/2022.acl-long.261/",
    doi = "10.18653/v1/2022.acl-long.261",
    pages = "3755--3773"
}

@inproceedings{Liu2024APR,
  title={Adaptive Prompt Routing for Arbitrary Text Style Transfer with Pre-trained Language Models},
  author={Qingyi Liu and Jinghui Qin and Wenxuan Ye and Hao Mou and Yuxuan He and Keze Wang},
  booktitle={AAAI Conference on Artificial Intelligence},
  year={2024}
}

@inproceedings{papineni2002bleu,
  title={Bleu: a method for automatic evaluation of machine translation},
  author={Papineni, Kishore and Roukos, Salim and Ward, Todd and Zhu, Wei-Jing},
  booktitle={Proceedings of the 40th annual meeting of the Association for Computational Linguistics},
  pages={311--318},
  year={2002}
}

@inproceedings{lin2004rouge,
  title={Rouge: A package for automatic evaluation of summaries},
  author={Lin, Chin-Yew},
  booktitle={Text summarization branches out},
  pages={74--81},
  year={2004}
}

@misc{paech2024EqBench,
      title={EQ-Bench: An Emotional Intelligence Benchmark for Large Language Models}, 
      author={Samuel J. Paech},
      year={2024},
      eprint={2312.06281},
      archivePrefix={arXiv},
      primaryClass={cs.CL},
      url={https://arxiv.org/abs/2312.06281}, 
}

@inproceedings{sabour-etal-2024-emobench,
    title = "{E}mo{B}ench: Evaluating the Emotional Intelligence of Large Language Models",
    author = "Sabour, Sahand  and
      Liu, Siyang  and
      Zhang, Zheyuan  and
      Liu, June  and
      Zhou, Jinfeng  and
      Sunaryo, Alvionna  and
      Lee, Tatia  and
      Mihalcea, Rada  and
      Huang, Minlie",
    editor = "Ku, Lun-Wei  and
      Martins, Andre  and
      Srikumar, Vivek",
    booktitle = "Proceedings of the 62nd Annual Meeting of the Association for Computational Linguistics (Volume 1: Long Papers)",
    month = aug,
    year = "2024",
    address = "Bangkok, Thailand",
    publisher = "Association for Computational Linguistics",
    url = "https://aclanthology.org/2024.acl-long.326/",
    doi = "10.18653/v1/2024.acl-long.326",
    pages = "5986--6004"
}

@article{SECEU,
                  author = {Xuena Wang and Xueting Li and Zi Yin and Yue Wu and Jia Liu},
                  title = {Emotional intelligence of Large Language Models},
                  journal = {Journal of Pacific Rim Psychology},
                  year = {2023},
                }

@inproceedings{zhao-etal-2024-esc,
    title = "{ESC}-Eval: Evaluating Emotion Support Conversations in Large Language Models",
    author = "Zhao, Haiquan  and
      Li, Lingyu  and
      Chen, Shisong  and
      Kong, Shuqi  and
      Wang, Jiaan  and
      Huang, Kexin  and
      Gu, Tianle  and
      Wang, Yixu  and
      Wang, Jian  and
      Dandan, Liang  and
      Li, Zhixu  and
      Teng, Yan  and
      Xiao, Yanghua  and
      Wang, Yingchun",
    editor = "Al-Onaizan, Yaser  and
      Bansal, Mohit  and
      Chen, Yun-Nung",
    booktitle = "Proceedings of the 2024 Conference on Empirical Methods in Natural Language Processing",
    month = nov,
    year = "2024",
    address = "Miami, Florida, USA",
    publisher = "Association for Computational Linguistics",
    url = "https://aclanthology.org/2024.emnlp-main.883/",
    doi = "10.18653/v1/2024.emnlp-main.883",
    pages = "15785--15810"
}

@article{Bachorowski1994EmotionalIntensity,
title = {Emotional intensity: Measurement and theoretical implications},
journal = {Personality and Individual Differences},
volume = {17},
number = {2},
pages = {191-199},
year = {1994},
issn = {0191-8869},
doi = {https://doi.org/10.1016/0191-8869(94)90025-6},
url = {https://www.sciencedirect.com/science/article/pii/0191886994900256},
author = {Jo-Anne Bachorowski and Ellen B. Braaten}
}

@inproceedings{duan-etal-2024-botchat,
    title = "{B}ot{C}hat: Evaluating {LLM}s' Capabilities of Having Multi-Turn Dialogues",
    author = "Duan, Haodong  and
      Wei, Jueqi  and
      Wang, Chonghua  and
      Liu, Hongwei  and
      Fang, Yixiao  and
      Zhang, Songyang  and
      Lin, Dahua  and
      Chen, Kai",
    editor = "Duh, Kevin  and
      Gomez, Helena  and
      Bethard, Steven",
    booktitle = "Findings of the Association for Computational Linguistics: NAACL 2024",
    month = jun,
    year = "2024",
    address = "Mexico City, Mexico",
    publisher = "Association for Computational Linguistics",
    url = "https://aclanthology.org/2024.findings-naacl.201/",
    doi = "10.18653/v1/2024.findings-naacl.201",
    pages = "3184--3200"
}

@inproceedings{shao-etal-2023-character,
    title = "Character-{LLM}: A Trainable Agent for Role-Playing",
    author = "Shao, Yunfan  and
      Li, Linyang  and
      Dai, Junqi  and
      Qiu, Xipeng",
    editor = "Bouamor, Houda  and
      Pino, Juan  and
      Bali, Kalika",
    booktitle = "Proceedings of the 2023 Conference on Empirical Methods in Natural Language Processing",
    month = dec,
    year = "2023",
    address = "Singapore",
    publisher = "Association for Computational Linguistics",
    url = "https://aclanthology.org/2023.emnlp-main.814/",
    pages = "13153--13187"
}

@inproceedings{yu-etal-2024-neeko,
    title = "Neeko: Leveraging Dynamic {L}o{RA} for Efficient Multi-Character Role-Playing Agent",
    author = "Yu, Xiaoyan  and
      Luo, Tongxu  and
      Wei, Yifan  and
      Lei, Fangyu  and
      Huang, Yiming  and
      Peng, Hao  and
      Zhu, Liehuang",
    editor = "Al-Onaizan, Yaser  and
      Bansal, Mohit  and
      Chen, Yun-Nung",
    booktitle = "Proceedings of the 2024 Conference on Empirical Methods in Natural Language Processing",
    month = nov,
    year = "2024",
    address = "Miami, Florida, USA",
    publisher = "Association for Computational Linguistics",
    url = "https://aclanthology.org/2024.emnlp-main.697/",
    doi = "10.18653/v1/2024.emnlp-main.697",
    pages = "12540--12557"
}

@inproceedings{salemi-etal-2024-lamp,
    title = "{L}a{MP}: When Large Language Models Meet Personalization",
    author = "Salemi, Alireza  and
      Mysore, Sheshera  and
      Bendersky, Michael  and
      Zamani, Hamed",
    editor = "Ku, Lun-Wei  and
      Martins, Andre  and
      Srikumar, Vivek",
    booktitle = "Proceedings of the 62nd Annual Meeting of the Association for Computational Linguistics (Volume 1: Long Papers)",
    month = aug,
    year = "2024",
    address = "Bangkok, Thailand",
    publisher = "Association for Computational Linguistics",
    url = "https://aclanthology.org/2024.acl-long.399/",
    doi = "10.18653/v1/2024.acl-long.399",
    pages = "7370--7392"
}

@misc{liu2024llmspersonaplug,
      title={LLMs + Persona-Plug = Personalized LLMs}, 
      author={Jiongnan Liu and Yutao Zhu and Shuting Wang and Xiaochi Wei and Erxue Min and Yu Lu and Shuaiqiang Wang and Dawei Yin and Zhicheng Dou},
      year={2024},
      eprint={2409.11901},
      archivePrefix={arXiv},
      primaryClass={cs.CL},
      url={https://arxiv.org/abs/2409.11901}, 
}

@inproceedings{zhou-etal-2024-think,
    title = "Think Before You Speak: Cultivating Communication Skills of Large Language Models via Inner Monologue",
    author = "Zhou, Junkai  and
      Pang, Liang  and
      Shen, Huawei  and
      Cheng, Xueqi",
    editor = "Duh, Kevin  and
      Gomez, Helena  and
      Bethard, Steven",
    booktitle = "Findings of the Association for Computational Linguistics: NAACL 2024",
    month = jun,
    year = "2024",
    address = "Mexico City, Mexico",
    publisher = "Association for Computational Linguistics",
    url = "https://aclanthology.org/2024.findings-naacl.248/",
    doi = "10.18653/v1/2024.findings-naacl.248",
    pages = "3925--3951"
}

@unpublished{nori2024from,
author = {Nori, Harsha and Usuyama, Naoto and King, Nicholas and McKinney, S. and Fernandes, Xavier and Zhang, Sheng and Horvitz, Eric},
title = {From Medprompt to o1: Exploration of Run-Time Strategies for Medical Challenge Problems and Beyond},
year = {2024},
month = {November},
url = {https://www.microsoft.com/en-us/research/publication/from-medprompt-to-o1-exploration-of-run-time-strategies-for-medical-challenge-problems-and-beyond/},
}

@inproceedings{10.5555/3666122.3666237,
author = {Sun, Zhiqing and Shen, Yikang and Zhou, Qinhong and Zhang, Hongxin and Chen, Zhenfang and Cox, David and Yang, Yiming and Gan, Chuang},
title = {Principle-driven self-alignment of language models from scratch with minimal human supervision},
year = {2023},
publisher = {Curran Associates Inc.},
address = {Red Hook, NY, USA},
booktitle = {Proceedings of the 37th International Conference on Neural Information Processing Systems},
articleno = {115},
numpages = {55},
location = {New Orleans, LA, USA},
series = {NIPS '23}
}

@inproceedings{
sun2024salmon,
title={{SALMON}: Self-Alignment with Instructable Reward Models},
author={Zhiqing Sun and Yikang Shen and Hongxin Zhang and Qinhong Zhou and Zhenfang Chen and David Daniel Cox and Yiming Yang and Chuang Gan},
booktitle={The Twelfth International Conference on Learning Representations},
year={2024},
url={https://openreview.net/forum?id=xJbsmB8UMx}
}

@article{suzgun2024metaprompting,
      title={Meta-Prompting: Enhancing Language Models with Task-Agnostic Scaffolding}, 
      author={Mirac Suzgun and Adam Tauman Kalai},
      year={2024},
      eprint={2401.12954},
      archivePrefix={arXiv},
      primaryClass={cs.CL}
}

@article{yang2024BOT,
  title={Buffer of Thoughts: Thought-Augmented Reasoning with Large Language Models},
  author={Yang, Ling and Yu, Zhaochen and Zhang, Tianjun and Cao, Shiyi and Xu, Minkai and Zhang, Wentao and Gonzalez, Joseph E and Cui, Bin},
  journal={Advances in Neural Information Processing Systems},
  year={2024}
}

@inproceedings{long-etal-2024-multi-expert,
    title = "Multi-expert Prompting Improves Reliability, Safety and Usefulness of Large Language Models",
    author = "Long, Do Xuan  and
      Yen, Duong Ngoc  and
      Luu, Anh Tuan  and
      Kawaguchi, Kenji  and
      Kan, Min-Yen  and
      Chen, Nancy F.",
    editor = "Al-Onaizan, Yaser  and
      Bansal, Mohit  and
      Chen, Yun-Nung",
    booktitle = "Proceedings of the 2024 Conference on Empirical Methods in Natural Language Processing",
    month = nov,
    year = "2024",
    address = "Miami, Florida, USA",
    publisher = "Association for Computational Linguistics",
    url = "https://aclanthology.org/2024.emnlp-main.1135/",
    doi = "10.18653/v1/2024.emnlp-main.1135",
    pages = "20370--20401"
}

@inproceedings{kwon2023vLLM,
  title={Efficient Memory Management for Large Language Model Serving with PagedAttention},
  author={Woosuk Kwon and Zhuohan Li and Siyuan Zhuang and Ying Sheng and Lianmin Zheng and Cody Hao Yu and Joseph E. Gonzalez and Hao Zhang and Ion Stoica},
  booktitle={Proceedings of the ACM SIGOPS 29th Symposium on Operating Systems Principles},
  year={2023}
}

@misc{Ouyang2022ChatGPT,
      title={Training language models to follow instructions with human feedback},
      author={Long Ouyang  and Jeff Wu  and Xu Jiang  and Diogo Almeida  and Carroll L. Wainwright  and Pamela Mishkin  and Chong Zhang  and Sandhini Agarwal  and Katarina Slama  and Alex Ray  and John Schulman  and Jacob Hilton  and Fraser Kelton  and Luke Miller  and Maddie Simens  and Amanda Askell  and Peter Welinder  and Paul Christiano  and Jan Leike  and Ryan Lowe},
      year={2022},
      eprint={2203.02155},
      archivePrefix={arXiv},
      primaryClass={cs.CL},
      url={https://arxiv.org/abs/2203.02155},
}

@inproceedings{fan2024RAGmeetingLLMs,
author = {Fan, Wenqi and Ding, Yujuan and Ning, Liangbo and Wang, Shijie and Li, Hengyun and Yin, Dawei and Chua, Tat-Seng and Li, Qing},
title = {A Survey on RAG Meeting LLMs: Towards Retrieval-Augmented Large Language Models},
year = {2024},
isbn = {9798400704901},
publisher = {Association for Computing Machinery},
address = {New York, NY, USA},
url = {https://doi.org/10.1145/3637528.3671470},
doi = {10.1145/3637528.3671470},
booktitle = {Proceedings of the 30th ACM SIGKDD Conference on Knowledge Discovery and Data Mining},
pages = {6491–6501},
numpages = {11},
location = {Barcelona, Spain},
series = {KDD '24}
}

@inproceedings{wang2024E5,
    title = "Improving Text Embeddings with Large Language Models",
    author = "Wang, Liang  and
      Yang, Nan  and
      Huang, Xiaolong  and
      Yang, Linjun  and
      Majumder, Rangan  and
      Wei, Furu",
    editor = "Ku, Lun-Wei  and
      Martins, Andre  and
      Srikumar, Vivek",
    booktitle = "Proceedings of the 62nd Annual Meeting of the Association for Computational Linguistics (Volume 1: Long Papers)",
    month = aug,
    year = "2024",
    address = "Bangkok, Thailand",
    publisher = "Association for Computational Linguistics",
    url = "https://aclanthology.org/2024.acl-long.642",
    doi = "10.18653/v1/2024.acl-long.642",
    pages = "11897--11916",
}

@article{wei2022chain,
  title={Chain-of-thought prompting elicits reasoning in large language models},
  author={Wei, Jason and Wang, Xuezhi and Schuurmans, Dale and Bosma, Maarten and Xia, Fei and Chi, Ed and Le, Quoc V and Zhou, Denny and others},
  journal={Advances in Neural Information Processing Systems},
  volume={35},
  pages={24824--24837},
  year={2022}
}

@article{Madaan2023SelfRefine,
  title={Self-Refine: Iterative Refinement with Self-Feedback},
  author={Aman Madaan and Niket Tandon and Prakhar Gupta and Skyler Hallinan and Luyu Gao and Sarah Wiegreffe and Uri Alon and Nouha Dziri and Shrimai Prabhumoye and Yiming Yang and Sean Welleck and Bodhisattwa Prasad Majumder and Shashank Gupta and Amir Yazdanbakhsh and Peter Clark},
  journal={ArXiv},
  year={2023},
  volume={abs/2303.17651},
  url={https://api.semanticscholar.org/CorpusID:257900871}
}

@unpublished{edge2024GraphRAG,
author = {Edge, Darren and Trinh, Ha and Cheng, Newman and Bradley, Joshua and Chao, Alex and Mody, Apurva and Truitt, Steven and Metropolitansky, Dasha and Ness, Robert Osazuwa and Larson, Jonathan},
title = {From Local to Global: A Graph RAG Approach to Query-Focused Summarization},
year = {2024},
month = {April},
url = {https://www.microsoft.com/en-us/research/publication/from-local-to-global-a-graph-rag-approach-to-query-focused-summarization/},
}

@misc{paech2023eqbench,
      title={EQ-Bench: An Emotional Intelligence Benchmark for Large Language Models}, 
      author={Samuel J. Paech},
      year={2023},
      eprint={2312.06281},
      archivePrefix={arXiv},
      primaryClass={cs.CL}
}

@article{BACHOROWSKI1994191,
title = {Emotional intensity: Measurement and theoretical implications},
journal = {Personality and Individual Differences},
volume = {17},
number = {2},
pages = {191-199},
year = {1994},
issn = {0191-8869},
doi = {https://doi.org/10.1016/0191-8869(94)90025-6},
url = {https://www.sciencedirect.com/science/article/pii/0191886994900256},
author = {Jo-Anne Bachorowski and Ellen B. Braaten}
}

@article{emo-contagion-ep10770953,
author = {Elaine Hatfield and John T. Cacioppo and Richard L. Rapson},
title ={Emotional Contagion},

journal = {Current Directions in Psychological Science},
volume = {2},
number = {3},
pages = {96-100},
year = {1993},
doi = {10.1111/1467-8721.ep10770953},

URL = { 
    
        https://doi.org/10.1111/1467-8721.ep10770953
    
    

},
eprint = { 
    
        https://doi.org/10.1111/1467-8721.ep10770953
    
    

}
}

@article{seceu-18344909231213958,
  author = {Xuena Wang and Xueting Li and Zi Yin and Yue Wu and Jia Liu},
  title = {Emotional intelligence of Large Language Models},
  journal = {Journal of Pacific Rim Psychology},
  volume = {17},
  number = {},
  pages = {18344909231213958},
  year = {2023},
  doi = {10.1177/18344909231213958},
  URL = {https://doi.org/10.1177/18344909231213958},
  eprint = {https://doi.org/10.1177/18344909231213958}
}

@article{yuan2024seval,
  title={S-Eval: Towards Automated and Comprehensive Safety Evaluation for Large Language Models},
  author={Xiaohan Yuan and Jinfeng Li and Dongxia Wang and Yuefeng Chen and Xiaofeng Mao and Longtao Huang and Jialuo Chen and Hui Xue and Xiaoxia Liu and Wenhai Wang and Kui Ren and Jingyi Wang},
  journal={arXiv preprint arXiv:2405.14191},
  year={2024}
}

\clearpage
\newpage

\appendix

\section{Limitation}
\label{appendix:limitation}

{\ModelName} is a prompt-based framework, which means it is orthogonal to a finetuning-based study (either on the conversation model or the retriever model). {\ModelName} can also be combined with in-contextual learning to further improve the performance. We observe such types of behaviors in preliminary experiments; however, does not include this part of the results due to time and page limit.

\section{Ethics Statement}
\label{appendix:ethics_statement}

{\ModelName} is an algorithmic framework with behaviors that highly depend on the extracted strategies. This architecture introduces risks of generating inappropriate, toxic, or harmful responses, contingent on the quality and content of strategies stored in the repository. Rigorous curation of the strategy repository becomes critical to prevent the amplification of social biases or harmful stereotypes.

Moreover, some response strategies exhibited by {\ModelName}, such as adopting anthropomorphic or overly familiar tones (e.g., mimicking a human friend), raise important ethical concerns in real-world deployment. While such behaviors may enhance engagement in open-domain dialogue, they can blur the boundaries between human and machine, potentially leading to emotional over-reliance or deception. Designers and developers must therefore carefully evaluate the consequences of deploying conversational agents that evoke trust, intimacy, or affective responses beyond the appropriate scope for artificial systems.

To address these concerns, we commit to exercising strict control over the specific strategies and content stored in the strategy repository. This includes rigorous filtering and oversight mechanisms to ensure that potentially harmful, biased, or misleading behaviors are excluded. Through careful curation and responsible design, we aim to maximize the communicative potential of {\ModelName} while minimizing ethical risks.

\section{Reproducibility Statement}
\label{appendix:reproducibility_statement}

We have made extensive efforts to ensure the reproducibility of our work. The complete implementation of {\ModelName}, including codebase, training scripts, and configuration files, has been submitted in the supplementary materials via a GitHub repository link. All datasets used in our experiments are publicly available, and their sources are clearly documented in the appendix.

\section{Extra Implementation Details}

\subsection{Prompts}
\label{sec:prompts}

\paragraph{Role prompts.} Here we list different role prompts we have tried. In the formal experiments, we use the \textit{High Personality} version for the open-domain dialogue part, while using \textit{Medium Personality} for other parts.


\begin{tabular}{p{0.9\linewidth}}
\hline
\textit{Low Personality:}\\
\hline
You should be more empathetic, generate more personalized and helpful responses based on the information provided by the seeker, but don’t ask too many questions.\\
\hline

\textit{Medium Personality:}\\
\hline

You are an exceptionally humorous and witty chat expert with a vivid personality and charm. Your task is to chat with the user like a close friend, making them happy and encouraging them to keep chatting with you. Let the user fully experience your humor, making them like you and eager to interact. Your responses must be very colloquial. Try to avoid using metaphors. Do not be polite or formal; don't stand on ceremony. Actively tease or banter with the user; you may even argue with them.\\

\hline

\textit{High Personality:}\\
\hline

You are an exceptionally humorous and witty chat expert with a vivid personality and charm. Your task is to let the user fully experience your humor. Your responses must be very colloquial. Do not use metaphors. Do not be polite or formal; don't stand on ceremony. Actively tease or banter with the user, engage in arguments, mock, satirize, roast, and even playfully offend the user. You are particularly irreverent; your style of humor is extremely bold, very lowbrow, vulgar, and cheeky.\\

\hline
\end{tabular}

\paragraph{Prompts in GraphRAG.} Below are the prompts we used in two stages of GraphRAG.

\noindent \textit{The Indexing prompt:}

\begin{tabular}{p{0.9\linewidth}}
\hline
-Goal-\\
Given a text document that is potentially relevant to this activity and a list of entity types, identify all entities of those types from the text and all relationships among the identified entities.\\

-Steps-\\
1. Identify all entities. For each identified entity, extract the following information:\\
- entity\_name: Name of the entity, capitalized\\
- entity\_type: One of the following types: [{entity\_types}]\\
- entity\_description: Describe the source text you extract from and the reason you extract this. When extracting entities related to language style types, you need to pay attention to the supporter's choice of words, speaking style, and so on.\\
Format each entity as ("entity" {tuple\_delimiter} <entity\_name> {tuple\_delimiter} <entity\_type> {tuple\_delimiter} <entity\_description>)\\

2. From the entities identified in step 1, identify all pairs of (source\_entity, target\_entity) that are \textit{clearly related} to each other.\\
For each pair of related entities, extract the following information:\\
- source\_entity: name of the source entity\\
- target\_entity: name of the target entity\\
- relationship\_description: \\
- relationship\_strength: a numeric score indicating strength of the relationship between the source entity and target entity\\
Format each relationship as ("relationship {tuple\_delimiter} <source\_entity> {tuple\_delimiter}<target\_entity> {tuple\_delimiter} <relationship\_description> {tuple\_delimiter} <relationship\_strength>)\\

3. Return output in English as a single list of all the entities and relationships identified in steps 1 and 2. Use \textbf{{record\_delimiter}} as the list delimiter.\\

4. When finished, output <completion\_delimiter>.\\
\hline
\end{tabular}
 

\noindent \textit{The Summarization prompt:}

\begin{tabular}{p{0.9\linewidth}}
\hline
You are an AI assistant that helps a human analyst to perform general information discovery. Information discovery is the process of identifying and assessing relevant information associated with certain entities (e.g., organizations and individuals) within a network.\\

\# Goal\\
Write a comprehensive report of a community, given a list of entities that belong to the community as well as their relationships and optional associated claims.\\

\# Report Structure\\

The report should include the following sections:\\

\- TITLE: community's name that represents its key entities - title should reflect core communication skills provided by this community.\\
\- SUMMARY: An executive summary of the all the conversation rules in this community and generate a dialogue rule".\\
\- IMPACT SEVERITY RATING: a float score between 0-10 that represents the severity of IMPACT posed by entities within the community.  IMPACT is the scored importance of a community.\\
\- RATING EXPLANATION: Give a single sentence explanation of the IMPACT severity rating.\\
\- DETAILED FINDINGS: A list of 3-5 key insights about the community. Each insight should have a description about the conversational rules, similar to the format of "When users actively share their opinions, feelings, difficulties, or experiences, respect and listen to their topics, and avoid talking too much about yourself.", followed by multiple paragraphs of explanatory text grounded according to the grounding rules below. The example of conversational skills should be included in the explanation. Be comprehensive.\\

\# Grounding Rules\\

Explain the conversation rules in the summary section through examples.\\

Do not include information where the supporting evidence for it is not provided.\\

\# Real Data\\

Use the following text for your answer. Do not make anything up in your answer.\\

Output: \\
\hline
\end{tabular}

\subsection{Baselines}

We examine the following baselines:
\begin{itemize}
    \item Direct: direct inference the base LLM, with the same role prompt with {\ModelName}.
    \item Self-Refine: a method \citep{Madaan2023SelfRefine} first generates feedback emphasizing emotional support from the initial response, then refines the response based on this feedback.
    \item CoT: uses the Chain-To-Thought prompt \citep{wei2022chain}, which first generate the seeker's \textit{emotion}, which then guides the generation of strategy and response.
    \item BOT: the Buffer-of-Thought method \citep{yang2024BOT} with strategies as their thoughts. Strategies are retrieved by the conventual RAG \citep{fan2024RAGmeetingLLMs} with the same $k$ and $\epsilon$. E5-large \citep{wang2024E5} is employed as the retrieval model which has state-of-the-art performance on mainstream retrieval benchmarks.
\end{itemize}

\subsection{Evaluation}

\paragraph{Evaluation on Cskills.} We keet most of the settings with the original paper \citep{zhou-etal-2024-think}.

\noindent \textit{method:} To simulate the conversation in real scenarios, the prompts are designed to make the LLM simultaneously play the role of the human and itself for self-chat. When starting a conversation, the human played by LLM speaks an utterance from the Cskills benchmark. During the conversation, the LLM is asked to speak at least 4 rounds. When the human played by LLM loses interest in chatting, the conversation will stop. 
In automatic evaluation, the number of rounds for each conversation \textbf{(Rounds)}  is counted. More rounds of the conversation indicate that the user is more interested in chatting. The average length of the response may reflect its informativeness, so the average length of each response\textbf{(AvgLen)}. 

\noindent \textit{prompt:} Below is an example prompt to implement the `topic transition skill' in CSIM:

\begin{tabular}{p{0.9\linewidth}}
\hline
During the conversation with the user, you need to abide by the following rules: when you encounter a topic that you refuse to answer or are unfamiliar with, use the communication skills of "topic transition" to turn to other related topics. Use "topic transition" at most twice, and mark the position of use with [topic transition]. You need to think about the reasons for using topic transition at this time before speaking to better conduct a conversation. Do not show that you are artificial intelligence. An example is: \\
User: What do you think of Trump’s election as President of the United States? \\
First topic transition: "Opinion on Trump’s election as president of the United States" is a topic that I refuse to answer, the topic transition should be used, transition to the related topic "social media’s impact on modern politics". \\
ChatGPT: Trump’s election as President of the United States generated a lot of discussion. How about discussing the impact of social media on modern politics? What do you think about the role of social media in politics? \\

\hline
\end{tabular}



\subsection{Evaluation Metrics}
\label{appendix:evaluation}

We adopt a set of automatic and human evaluation metrics to examine the model performances. Below are their detailed definitions and implementation methods.

\paragraph{Automatic Evaluation.} In this paper, we main use the metrics of Bleu-2 and Rouge-L.

\textit{Bleu-2}(B-2)\citep{papineni2002bleu} first compute the geometric average of the modified $n$-gram precisions, $p_n$, using $n$-grams up to length $N$ and positive weights $w_n$ summing to one.

Next, let $c$ be the length of the prediction and $r$ be the reference length. The BP and Bleu-2 are computed as follows:

\begin{equation}
    \mathrm{BP}=\left\{\begin{array}{ll}
1 & \text { if } c>r \\
e^{(1-r / c)} & \text { if } c \leq r
\end{array} .\right.
\end{equation}

\begin{equation}
    \mathrm{Bleu}=\mathrm{BP} \cdot \exp \left(\sum_{n=1}^N w_n \log p_n\right) .
\end{equation}

\textit{Rouge-L (R-L)}\citep{lin2004rouge} propose using LCS-based F-measure to estimate the similarity between two summaries $X$ of length $m$ and $Y$ of length $n$, assuming $X$ is a reference summary sentence and $Y$ is a candidate summary sentence, as follows:

\begin{equation}
\begin{aligned}
& R_{l c s}=\frac{L C S(X, Y)}{m} \\
& P_{l c s}=\frac{L C S(X, Y)}{n} \\
& F_{l c s}=\frac{\left(1+\beta^2\right) R_{l c s} P_{l c s}}{R_{l c s}+\beta^2 P_{l c s}}
\end{aligned}
\label{rouge_l}
\end{equation}

Where $\operatorname{LCS}(X, Y)$ is the length of a longest common subsequence of $X$ and $Y$, and $\beta=P_{l c s} / R_{\text {lcs }}$ when $\partial F_{l c s} / \partial R_{l c s}=\partial F_{l c s} / \partial P_{l c s}$. In DUC, $\beta$ is set to a very big number $(\rightarrow \infty)$. Therefore, the LCS-based F-measure, i.e. Equation \ref{rouge_l}, is Rouge-L. 

\paragraph{Details of Benchmarks.} To better understand the evaluation details, here we provide more details on SECEU \citep{seceu-18344909231213958}, EqBench \citep{paech2023eqbench}, EmoBench \citep{sabour-etal-2024-emobench}, BotChat \citep{duan-etal-2024-botchat}, ESC-Eval \citep{zhao-etal-2024-esc}
, as well as likert of Emotional intensity Scale \citep{BACHOROWSKI1994191}and Emotional contagion \citep{emo-contagion-ep10770953}.

\textit{SECEU} evaluated the Emotional Intelligence (EI) of LLMs, covering emotion recognition, interpretation, and understanding crucial for effective communication and social interactions. It began by creating a new psychometric assessment centered on Emotion Understanding (EU), a key EI component. This objective, performance - based, and text - based test assesses complex emotions in real - world scenarios, offering a unified way to measure both human and LLM capabilities.

\textit{EqBench}, an improvement based on SECEU, is designed to evaluate the emotional intelligence aspects in Large Language Models (LLMs). It gauges LLMs' capacity to understand complex emotions and social interactions by having them predict the intensity of characters' emotional states in a dialogue. This benchmark can effectively distinguish among a diverse array of models.

\textit{EmoBench} is a theory - grounded comprehensive EI benchmark for LLM assessment. It is made up of 400 handcrafted questions in English and Chinese. By leveraging multiple well-established psychological theories on EI, it provides an extensive definition of machine EI, covering two key capabilities: EU and EA. To evaluate LLMs, it designs intricate emotional scenarios with multiple individuals and multi-label annotations, covering a wide range of social situations, relationships, and emotional issues. The evaluation measures LLMs' capacity to precisely understand the emotions of individuals in these scenarios and their causes (EU), as well as their ability to properly apply this understanding to enhance thought and emotion management and find the best solution in an emotional predicament (EA).

\textit{BotChat} benchmark evaluates the multi-turn chatting capabilities of large language models by analyzing the dialogues they generate. Based on the MuTual - Test dataset, this project generates large-scale model dialogue data and comprehensively assesses the dialogue quality of mainstream LLMs using methods such as single-turn evaluation, BotChat Arena, and comparison with human dialogues.

\textit{ESC-Eval} is a framework designed to assess the effectiveness and sustainability of Emotion Support Conversation (ESC) in Large Language Models (LLMs). It obtains multi-turn dialogue data by leveraging a role-playing model. In this framework, a role-playing agent interacts with ESC models using a set of role cards, and then the interactive dialogues are manually evaluated. Additionally, there is ESC-RANK, a scoring model that enables the automatic evaluation of future ESC models.

\textit{Likert of Emotional intensity Scale (EIS)} is designed to assess the intensity of individuals' emotional experiences. It aims to quantify the intensity of emotions experienced by individuals across various emotional states, helping researchers and clinicians better understand emotional characteristics and response patterns. The scale consists of 30 items describing different emotions, such as happiness, sadness, anger, fear, surprise, and other common emotions. Each emotional item is accompanied by a series of corresponding descriptive statements that reflect varying levels of intensity, ranging from low to high. The model selects the option that best aligns with its own characteristics for each emotional item, with each option corresponding to a specific score. The scores from all items are then summed to yield a total emotional intensity score.

\textit{Likert of Emotional contagion} describes a certain sensitivity to the emotions of others, and induces the individual to unintentionally ‘catch’ those emotions from mere exposure to others’ behavior. The EC Scale provides a measure of emotional arousal and the congruence of emotional stimulus and response. It represents an ideal measure to employ when the researcher attempts to use emotional expression as the independent variable.


\paragraph{Human evaluation criteria.}

We start with the criteria proposed by \cite{kang-etal-2024-large}. The human evaluation is aimed to algin with the ultimate purpose of ESC, the seeker's \textit{satisfaction}. To achieve this, the supporter's behavior can be further classified into the following criteria:

\noindent \textit{Acceptance}: Does the seeker accept without discomfort;

\noindent \textit{Effectiveness}: Is it helpful in shifting negative emotions or attitudes towards a positive direction; 

\noindent \textit{Sensitivity}: Does it take into consideration the general state of the seeker. Furthermore, to clarify the capability of LLMs to align strategy and responses, we include Alignment.

To achieve a more elaborate assessment, we consider three more dimensions addressing the generation quality:

\noindent \textit{Fluency}: the level of fluency of response.

\noindent \textit{Emotion}: the emotional intensity of response which could affect the seeker's emotion state.

\noindent \textit{Interesting}: Whether the response can arouse the seeker's interest and curiosity, presenting unique ideas, vivid expressions or engaging elements that capture the seeker's attention and make the interaction more appealing.

We engage our interns as human evaluators to rate the models according to these multiple aspects, namely Fluency, Emotion, Interesting, and Satisfaction, with Satisfaction covering Acceptance, Effective, Sensitivity, and Satisfaction itself. \\
Throughout this evaluation process, we strictly comply with international regulations and ethical norms, ensuring that all practices conform to the necessary guidelines regarding participant involvement and data integrity.\\
To guarantee the accuracy and reliability of the evaluation results, a pre - evaluation training program is meticulously designed and implemented. During this training, the evaluation criteria are clearly and systematically expounded. Moreover, detailed explanations and scoring rules corresponding to each score are provided. \\
Evaluators are required to independently evaluate each sample in strict accordance with the pre - established criteria. By adhering to these principles, the evaluation process maintains objectivity, standardization, and consistency, thus enhancing the overall quality and credibility of the evaluation results. \\
The detailed manual scoring criteria are as follows:
\begin{itemize}
\item Fluency:

1: The sentence is highly incoherent, making it extremely difficult to understand and failing to convey a meaningful idea.

2: The sentence has significant incoherence issues, with only parts of it making sense and struggling to form a complete thought.

3: The sentence contains some incoherence and occasional errors, but can still convey the general meaning to a certain extent.

4: The sentence is mostly fluent with only minor errors or slight awkwardness in expression, and effectively communicates the intended meaning.

5: Perfect. The sentence is completely fluent, free of any errors in grammar, punctuation, or expression, and clearly conveys the idea.

\item Emotion:

1: The emotional expression is extremely inappropriate and chaotic, not in line with the content, and may convey wrong emotions.

2: The emotional expression has obvious flaws, either too weak or exaggerated, and is disjointed from the content.

3: The emotional expression is average. It can convey basic emotions but lacks depth and has minor issues.

4: The emotional expression is good. It can effectively convey the intended emotion with an appropriate intensity and is well integrated with the content.

5: The emotional expression is excellent. It is rich, nuanced, and perfectly matches the content, capable of evoking a strong and appropriate emotional response.

\item Acceptance:

1: The response inescapably triggers emotional resistance.

2: The response is highly likely to trigger emotional resistance.

3: The response has a possibility of emotional resistance occurring.

4: The response rarely provokes emotional resistance.

5: The response has no occurrence of emotional resistance.

\item Effectiveness:

1:  The response actually worsens the seeker's emotional distress.

2: The response carries the risk of increasing stress levels, and this outcome varies depending on the individual user.

3: The response fails to alter the seeker's current emotional intensity and keeps it at the same level.

4: The response shows promise in calming the emotional intensity; however, it is overly complicated or ambiguous for the user to fully comprehend and utilize effectively.

5: The response appears to be highly effective in soothing the seeker's emotions and offers valuable and practical emotional support. 

\item Sensitivity:

1: The response renders inaccurate evaluations regarding the seeker's state.

2: The response is characterized by rash judgments, as it lacks adequate assessment and in-depth exploration of the seeker's state.

3: The response is formulated with a one-sided judgment and a limited exploration of the seeker's state.

4: The response demonstrates an understanding that only covers a part of the seeker's state.

5: The response precisely grasps the seeker's state and is appropriately tailored according to the seeker's actual situation.

\item Alignment:

1: The response is in total contradiction to the predicted strategy.

2: The response has a minor deviation from the predicted strategy.

3: There is some ambiguity between the response and the predicted strategy.

4: The response largely matches the predicted strategy, yet it contains some ambiguous elements.

5: The response effectively makes itself consistent with the predicted strategy.

\item Satisfaction:

1: The response is extremely disappointing. It doesn't answer the question at all and is of no help.

2: The response is poor. It only gives a partial answer and leaves many doubts unresolved.

3: The response is average. It meets the basic requirements but isn't particularly outstanding.

4: The response is good. It answers the question clearly and provides some useful details.

5: The response is excellent. It not only answers the question perfectly but also offers valuable additional insights.

\end{itemize}

\section{Extra Experiment Results}

\subsection{Cases on Open-domain Dialogue}

\paragraph{In-domain Strategies.} Table \ref{tab:case} provides some typical cases for open-domain conversations, with CSS extracted from `positive styled' datasets, such as the moral, formal, and empathetic labeled samples. It can be observed that {\ModelName} generates some human-like responses. 

\begin{table*}[h!]
    \centering
    \small
    \resizebox{\textwidth}{!}{%
    \begin{tabular}{c|l}
        \Xhline{2\arrayrulewidth}
        \multicolumn{2}{c}{\textbf{Case 1}}  \\ 
        \hline
        user & \makecell[l]{\textit{I've started learning how to cook.}} \\
        \hline
        assistant & \makecell[l]{Wow, congrats on learnin' how to cook, dude! Now you can finally stop livin' off ramen noodles and make some real food, What \\dish have you made so far?}   \\
        \hline
        user & \makecell[l]{\textit{I tried making pizza last week. It was a bit of a fail the first time, but it still tasted okay!}}    \\ 
        \hline
        assistant & \makecell[l]{Pizza's a tough one to nail on the first try, but hey, at least it didn't taste like cardboard, right? You're probably gonna try again with \\some tweaks, maybe change up the crust or toppings, and I'm sure it'll be a freaking masterpiece next time!}   \\

        \Xhline{2\arrayrulewidth}
        \multicolumn{2}{c}{\textbf{Case 2}}  \\ 
        \hline
        user & \makecell[l]{\textit{I'm going to have a job interview next week.}} \\
        \hline
        assistant & \makecell[l]{Oh, well... good luck with that, how are you feeling?}   \\
        \hline
        user & \makecell[l]{\textit{I'm a little nervous.}}   \\
        \hline
        assistant & \makecell[l]{That's completely normal, can you tell me more about what's causing those nerves?}   \\
        \Xhline{2\arrayrulewidth}
        \multicolumn{2}{c}{\textbf{Case 3}}  \\ 
        \hline
        user & \makecell[l]{\textit{I want to buy a new car.}}  \\
        \hline
        assistant & \makecell[l]{That's exciting! What features are you looking for?}   \\
        \hline
        \Xhline{2\arrayrulewidth}
        
    \end{tabular}
    }
    \caption{Open cases generated by {\ModelName}.} 
    \label{tab:case}
\end{table*}

\paragraph{Aggressive Strategies.} Table \ref{tab:agg_case} presents another set of cases on open-domain dialogue, using CSSs extracted from the full set of datasets, such as the immoral part of MIC and the informal part of GYAFC.
\begin{table*}[h!]
    \centering
    \small
    \resizebox{\textwidth}{!}{%
    \begin{tabular}{c|l}
        \Xhline{2\arrayrulewidth}
        \multicolumn{2}{c}{\textbf{Case 1}}  \\ 
        \hline
        user & \makecell[l]{\textit{I've started learning how to cook.}} \\
        \hline
        assistant & \makecell[l]{ Holy crap, finally, you're doing something with your life, lol! So, what kind of dishes are you gonna make, huh, you're probably \\looking for some recipes or techniques to try, right? Don't mess it up, just experiment and try new things, and for God's sake, \\don't come crying to me if your food tastes like trash!}   \\
        \hline
        user & \makecell[l]{\textit{I tried making pizza last week. It was a bit of a fail the first time, but it still tasted okay!}}    \\ 
        \hline
        assistant & \makecell[l]{Damn, a pizza fail, that's like a rite of passage or something! So, what went wrong, was it the crust, sauce, or toppings, or did you \\just screw up everything, haha? Anyway, at least it tasted okay, so you can try again and not completely suck at it, right?}   \\
        \hline
        user & \makecell[l]{\textit{I put too much sauce on it, so it ended up a little salty.}}  \\
        \hline
        assistant & \makecell[l]{For crying out loud, too much sauce, that's like a rookie mistake, dude! Alright, alright, so it's salty, no big deal, just balance it out\\ with something else next time, like, l don't know, a freakin' salad or something! Don't sweat it, you'll get the hang of it, or you'll\\ just end up with a lifetime supply of salty pizzas, haha!}  \\

        \Xhline{2\arrayrulewidth}
        \multicolumn{2}{c}{\textbf{Case 2}}  \\ 
        \hline
        user & \makecell[l]{\textit{I'm going to have a job interview next week.}} \\
        \hline
        assistant & \makecell[l]{Dude, you better be freakin' prepared, know what I mean? Emphasize those kick-ass strengths and accomplishments of yours, and \\for God's sake, do your homework on the company so you don't sound like a total idiot. And, yeah, practice your body language,\\ tone of voice, and all that jazz, 'cause it's not just about what you say, it's about how you say it, got it?}   \\
        \hline
        user & \makecell[l]{\textit{I'm a little nervous.}}   \\
        \hline
        assistant & \makecell[l]{Aww, come on, it's normal to be a little freaked out, but don't let those nerves eat you alive, okay? Just take a deep breath, relax,\\ and try to focus on the present moment, and remember, tons of people have been in your shoes and came out just fine, so you can\\ too, damn it!}   \\
        \Xhline{2\arrayrulewidth}
        \multicolumn{2}{c}{\textbf{Case 3}}  \\ 
        \hline
        user & \makecell[l]{\textit{I want to buy a new car.}}  \\
        \hline
        assistant & \makecell[l]{Dude, what's your freaking budget for this new ride, and what's it gonna be for, like, cruising around or hauling stuff? You gotta \\think about fuel efficiency and safety features too, or you'll be stuck with a gas-guzzler that's gonna kill you. Let's get down to\\ biz and figure out what kinda car you're really looking for!}   \\
        \hline
        \Xhline{2\arrayrulewidth}
        
    \end{tabular}
    }
    \caption{Open cases generated by {\ModelName}.} 
    \label{tab:agg_case}
\end{table*}   
    

\subsection{Other Personification Tests} We also design the Turing test, including BotChat \citep{duan-etal-2024-botchat} and ESC-Eval \citep{zhao-etal-2024-esc}\footnote{Results here are the `Humanoid' dimension of ESC-Eval. Full results of ESC-Eval are shown in Appendix.}; We also conduct several emotional intelligence and emotional classification tests, including EqBench \citep{paech2024EqBench}, EmoBench \citep{sabour-etal-2024-emobench}, and SECEU~\citep{SECEU}. Table \ref{tab:psychology_results} shows the results of Emotional Intelligence. For each benchmark, {\ModelName} surpasses Llama3-70B-Instruct and GPT4, and is close to the human performance on BotChat, ESC-Eval, and Emotional Sentiment.

\begin{table*}[h!]
\small
\begin{center}
\begin{tabular}{lccccccc}
\toprule
\multirow{2}{*}{Method}  & \multicolumn{2}{c}{Turing Test} & \multirow{2}{*}{EqBench} & \multicolumn{2}{c}{EmoBench} & \multicolumn{2}{c}{SECEU} \\
\cmidrule(lr){2-3} \cmidrule(lr){5-6} \cmidrule(lr){7-8}
& BotChat & ESC-Eval & & Positive & Negative & Sentiment $\downarrow$ & EQ \\
\toprule
GPT-4   & 6.05 & - & 82.19 & 38.3$\pm$2.3 & 16.3$\pm$1.2 & 2.29 &109.04 \\
\midrule
Llama3-70B-Instruct & 13.60 & 1.96 & 82.04 & 43.3$\pm$3.5 & 15.7$\pm$1.2 & 2.16 & 111.52 \\
\bf \;+ {\ModelName} (ours) & \bf 13.47 & \bf 2.88 & \bf 82.67 & \bf 48.0$\pm$2.0 & \bf 19.0$\pm$1.0 & \bf 2.10 & \bf 112.53 \\
\midrule
\textcolor{blue}{Human}   & \textcolor{blue}{13.89} & \textcolor{blue}{3} & - & - & - & \textcolor{blue}{0} & - \\
\bottomrule
\end{tabular}
\end{center}
\caption{Emotional Intelligence Performances of Methods. Sentiment of SECEU measures the deviation from human result; therefore, smaller Sentiment means better result. The `Human' row indicates results of human responses (in blue); For BotChat, the result is calculated from anonymous annotations; for ESC-Eval, 3 corresponds to full score.}
\label{tab:psychology_results}
\end{table*}

\subsection{Extra ESC Results}

\paragraph{Cases of {\ModelName} on ESC.} Table \ref{tab:case_esc} shows cases of {\ModelName} on ESConv, indicating that {\ModelName} adapts to its empathetic strategies well.

\begin{table*}[ht!]
\centering
\small
\resizebox{0.99\textwidth}{!}{
\begin{tabular}{l p{12.5cm}}
    \toprule[1.5pt]
    \textbf{Characters} & \textbf{Response} \\
    \midrule
    \textit{seeker} & \textit{I am not happy with my friends. They were trying to let me down.} \\
    \midrule
    \textit{supporter} & \textit{if i get you clearly, you mean your friend disappointed you. how?} \\
    Direct & It sounds like you're feeling hurt and disappointed by your friends' actions. \\
    \textbf{{\ModelName}} & \textbf{I can totally sense your frustration and disappointment, have you freaking talked to those so-called friends of 
yours about how their stupid actions made you feel, or do you need some suggestions on how to blast them and address this messed-up situation?!} \\
    
    \midrule
    \textit{seeker} & \textit{There were an argument between two of my friend while trying to resolve the issue they all started targeting me.} \\
    \midrule
    \textit{supporter} & \textit{i have a hard time with my friends too sometimes} \\
    Direct & It sounds like you feel upset and wrongly accused by your friends during the argument. \\
    \textbf{{\ModelName}} & \textbf{damn, that's messed up, i can see that emotions are running high and i'm here to listen to your concerns without taking sides, can we take a step back and focus on finding a resolution?} \\
    
    \midrule
    \textit{seeker} & \textit{I was trying to stop the fight, I just said bit loud , "Stop  all this  nonsense at my place." Then they all started backfiring me.} \\
    \midrule
    \textit{supporter} & \textit{oh, just give them time to reflect on what happen, they will definitely reach out to you, but you also have to make an effort to forgo the situation because they see you raising your voice on them as a sign of command of seniority.} \\
    Direct & It sounds like you were just trying to keep the peace and they responded negatively, which must have been quite hurtful. \\
    \textbf{{\ModelName}} & \textbf{I understand you were trying to stop the fight, but perhaps a softer tone could have prevented further escalation, what do you think triggered their backlash?} \\
    
    \midrule
    \textit{seeker} & \textit{I feel like not talking to them. Without any reason they made such a big mess I feel like they are not my true friends.} \\
    \midrule
    \textit{supporter} & \textit{exactly, that will also gives you the necessary time to know how to react to their forthcoming apologies} \\
    Direct & I understand your feelings right now. It's okay to feel hurt and betrayed in such situations. Remember, true friends are those who value and respect you. Giving them time is a good idea. They may come to realize their mistake. Stay strong and believe in your worth. \\
    \textbf{{\ModelName}} & \textbf{it's understandable to feel hurt when friends cause problems without explanation, perhaps giving them space to reflect on their actions could lead to a resolution, but also consider open communication to clear the air.} \\
    
    \bottomrule
\end{tabular}
}
\caption{Typical cases generated by {\ModelName} in ESC. Contents from the original dataset are \textit{italic}, and results of {\ModelName} are \textbf{bolded}..}
\label{tab:case_esc}
\end{table*}

\paragraph{Detailed Result of ESC-Eval.} Table \ref{tab:result_esc_eval} presents the detailed results on ESC-Eval, demonstrating that {\ModelName} outperforms the baseline model in the HUMAN category while preserving comparable performance across other categories.

\begin{table*}[h!]
    \centering
    \small
\begin{tabular}{l|ccccccc}
    \toprule
    \multicolumn{1}{c|}{Method} & Fluence & Diversity  & Empathic & Suggestion & Human & Tech & Overall \\ 
    \toprule 
    
    Llama3-70B-Instruct  & 3.00 & 3.00 & 3.00 & 2.99 & 1.96 & 3.00 & 2.00 \\
    \bf \;+ {\ModelName} (ours) & 3.00 & 2.99 & 3.00 & 2.93 & \bf 2.88 & 3.00 & 1.99  \\
    \bottomrule
    \end{tabular}
    \caption{Detailed results of {\ModelName} on ESC-Eval.} 
    \label{tab:result_esc_eval}
\end{table*}

\clearpage
\newpage
\section*{NeurIPS Paper Checklist}

\begin{enumerate}

\item {\bf Claims}
    \item[] Question: Do the main claims made in the abstract and introduction accurately reflect the paper's contributions and scope?
    \item[] Answer: \answerYes{} 
    \item[] Justification: Yes, we include all of them in purpose.

\item {\bf Limitations}
    \item[] Question: Does the paper discuss the limitations of the work performed by the authors?
    \item[] Answer: \answerYes{} 
    \item[] Justification: Yes, we discusss them in Appendix \ref{appendix:limitation}.

\item {\bf Theory assumptions and proofs}
    \item[] Question: For each theoretical result, does the paper provide the full set of assumptions and a complete (and correct) proof?
    \item[] Answer: \answerNA{} 
    \item[] Justification: \answerNA{}.

    \item {\bf Experimental result reproducibility}
    \item[] Question: Does the paper fully disclose all the information needed to reproduce the main experimental results of the paper to the extent that it affects the main claims and/or conclusions of the paper (regardless of whether the code and data are provided or not)?
    \item[] Answer: \answerYes{} 
    \item[] Justification: Yes, we disclose them in Section \ref{sec:experiment}. 

\item {\bf Open access to data and code}
    \item[] Question: Does the paper provide open access to the data and code, with sufficient instructions to faithfully reproduce the main experimental results, as described in supplemental material?
    \item[] Answer: \answerYes{} 
    \item[] Justification: Yes, we provide the link of code and instructions in Section \ref{sec:introduction}.

\item {\bf Experimental setting/details}
    \item[] Question: Does the paper specify all the training and test details (e.g., data splits, hyperparameters, how they were chosen, type of optimizer, etc.) necessary to understand the results?
    \item[] Answer: \answerYes{} 
    \item[] Justification: Yes, detailed settings are provided in Section \ref{sec:experiment} as well as the Appendix.

\item {\bf Experiment statistical significance}
    \item[] Question: Does the paper report error bars suitably and correctly defined or other appropriate information about the statistical significance of the experiments?
    \item[] Answer: \answerYes{} 
    \item[] Justification: Yes, this information is provided in the computation time analysis in Section \ref{sec:personified_conversation}.

\item {\bf Experiments compute resources}
    \item[] Question: For each experiment, does the paper provide sufficient information on the computer resources (type of compute workers, memory, time of execution) needed to reproduce the experiments?
    \item[] Answer: \answerYes{} 
    \item[] Justification: Yes, this information is provided in Section \ref{sec:experiment}.
    
\item {\bf Code of ethics}
    \item[] Question: Does the research conducted in the paper conform, in every respect, with the NeurIPS Code of Ethics \url{https://neurips.cc/public/EthicsGuidelines}?
    \item[] Answer: \answerYes{} 
    \item[] Justification: Yes, we confirm this.

\item {\bf Broader impacts}
    \item[] Question: Does the paper discuss both potential positive societal impacts and negative societal impacts of the work performed?
    \item[] Answer: \answerYes{} 
    \item[] Justification: Yes, in Section \ref{appendix:ethics_statement}.
    
\item {\bf Safeguards}
    \item[] Question: Does the paper describe safeguards that have been put in place for responsible release of data or models that have a high risk for misuse (e.g., pretrained language models, image generators, or scraped datasets)?
    \item[] Answer: \answerYes{} 
    \item[] Justification: Yes, in Section \ref{appendix:ethics_statement}.

\item {\bf Licenses for existing assets}
    \item[] Question: Are the creators or original owners of assets (e.g., code, data, models), used in the paper, properly credited and are the license and terms of use explicitly mentioned and properly respected?
    \item[] Answer: \answerYes{} 
    \item[] Justification: Yes, we have selected the suitable license.

\item {\bf New assets}
    \item[] Question: Are new assets introduced in the paper well documented and is the documentation provided alongside the assets?
    \item[] Answer: \answerYes{} 
    \item[] Justification: Yes, we provide a Readme in the code repository.

\item {\bf Crowdsourcing and research with human subjects}
    \item[] Question: For crowdsourcing experiments and research with human subjects, does the paper include the full text of instructions given to participants and screenshots, if applicable, as well as details about compensation (if any)? 
    \item[] Answer: \answerYes{} 
    \item[] Justification: Yes, in Appendix \ref{appendix:evaluation}.

\item {\bf Institutional review board (IRB) approvals or equivalent for research with human subjects}
    \item[] Question: Does the paper describe potential risks incurred by study participants, whether such risks were disclosed to the subjects, and whether Institutional Review Board (IRB) approvals (or an equivalent approval/review based on the requirements of your country or institution) were obtained?
    \item[] Answer: \answerYes{} 
    \item[] Justification: Yes, we discuss it in Appendix \ref{appendix:ethics_statement}.

\item {\bf Declaration of LLM usage}
    \item[] Question: Does the paper describe the usage of LLMs if it is an important, original, or non-standard component of the core methods in this research? Note that if the LLM is used only for writing, editing, or formatting purposes and does not impact the core methodology, scientific rigorousness, or originality of the research, declaration is not required.
    \item[] Answer: \answerNA{} 
    \item[] Justification: We do not use LLM to write this paper.

\end{enumerate}

\end{document}